\documentclass[10pt,twocolumn,letterpaper]{article}
\usepackage{iccv}
\usepackage{times}
\usepackage{epsfig}
\usepackage{graphicx}
\usepackage{amsmath}
\usepackage{amssymb}
\usepackage{subfig}
\usepackage[font={small}]{caption}
 \iccvfinalcopy 
\def\iccvPaperID{2127} 

\begin{document}
\title{Robust Motion Segmentation from Pairwise Matches}
\author{Federica Arrigoni and Tomas Pajdla\\
\small Czech Institute of Informatics, Robotics and Cybernetics -- Czech Technical University in Prague \\
\small Jugoslavskych partyzanu 3, Prague, Czech Republic \\
{\tt\small Federica.Arrigoni@cvut.cz, pajdla@cvut.cz}
}
\maketitle
\newcommand{\TP}[1]{{\bf TP: #1}}
\newcommand{\FA}[1]{{\bf FA: #1}}
\newcommand{\ourmethodall}{\textsc{Mode-All}\xspace} 
\newcommand{\ourmethod}{\textsc{Mode}\xspace} 
\begin{abstract}
\noindent In this paper we address a classification problem that has not been considered before, namely motion segmentation given pairwise matches only. Our contribution to this unexplored task is 
a novel formulation of motion segmentation as a two-step process. First, motion segmentation is performed on image pairs independently. Secondly, we combine independent pairwise segmentation results in a robust way into the final globally consistent segmentation. Our approach is inspired by the success of averaging methods. 
We demonstrate in simulated as well as in real experiments that our method is very effective in reducing the errors in the pairwise motion segmentation and can cope with large number of mismatches.
\end{abstract}
\section{Introduction}
\noindent Motion segmentation is an essential task in many applications in Computer Vision and Robotics, such as surveillance~\cite{KimKim03}, action recognition~\cite{WeinlandRonfardAl11} and scene understanding~\cite{EssTobiasAl09}. The classic way to state the problem is the following: given a set of feature points that are \emph{tracked} through a sequence of images, the goal is to cluster those trajectories according to the different motions they belong to. It is assumed that the scene contains multiple objects that are moving rigidly and independently in 3-space.
There is a plenty of available techniques to accomplish such task, as detailed in Sec.~\ref{sec:relatedwork}. Among them, the methods developed in \cite{JiSalzmanLi15,LaiWangAl17,XuCheongAl18} achieve a very low misclassification error on the Hopkins155 benchmark \cite{TronVidal07}, which is a well established dataset to test the performance of motion segmentation. However, the tracks available in the dataset are not realistic at all since they were filtered with the aid of manual operations.
\begin{figure}[htbp] 
  \centering
\includegraphics[width=1\linewidth]{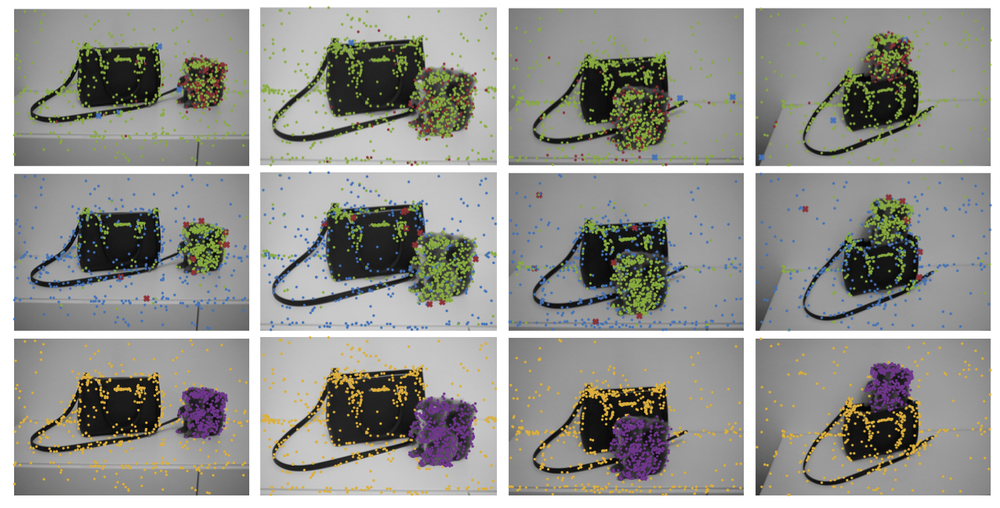}
\caption{
Segmentation results are reported on four images 
for the technique in \cite{XuCheongAl18} combined with \cite{OlssonEnqvist11} (top) and our method (middle). Image points are drawn in different colours: green (correctly classified); red (misclassified); blue (unknown). Ground-truth segmentation is also reported (bottom) where different colors encode the membership to different motions.
}
\label{fig:final_segmentation}
\end{figure}
In this paper we take motion segmentation one step further by considering more difficult/realistic assumptions, namely we assume that pairwise matches (e.g.~those computed from SIFT keypoints \cite{Lowe04}) are available only, and we address the task of classifying image points (instead of tracks), as shown in Fig.~\ref{fig:final_segmentation}.
This problem has not been considered before but it has great practical relevance since it does not require to compute tracks in advance, which is a challenging task in the presence of multiple moving objects. 

More precisely, we formulate motion segmentation as a two-step process:
\begin{enumerate}
    \item segmentation of corresponding points is performed on each image pair in isolation;
    \item segmentation of image points is computed without relying on the feature locations, using only the classification of matching points derived in Step~1.
\end{enumerate}
Our new formulation is detailed in Sec.~\ref{sec:formulation}. Regarding Step~2, we develop a simple scheme that classifies each point based on the frequencies of labels of that point in different image pairs, which is reported in Sec.~\ref{sec:our_method}. The resulting method is a general framework that can be combined with any algorithm able to perform motion segmentation in two images. 

The idea of combining results from individual image pairs was also present in \cite{LiGuoAl13}, where all the pairs were considered, and in~\cite{LaiWangAl17,XuCheongAl18}, where only pairs of consecutive frames were used. These techniques, however, are different from our approach since they do not completely perform segmentation of image pairs but they rely on \emph{partial} results only (i.e.~correlation of corresponding points).
Such results are used to build an affinity matrix that encodes the similarity between different tracks, to which spectral clustering \cite{Von-Luxburg07} (or its multi-view variations \cite{CortesMohriAl09,KumarRaiAl11,WangQianAl14}) is applied. As a consequence, they perform segmentation of tracks, whereas our method classifies image points. A related approach \cite{JiLiAl14} considers the scenario where correspondences are unknown and uses the Alternating Direction Method of Multipliers (ADMM) \cite{BoydParikh11} to jointly perform motion segmentation and tracking, where sequences with at most 200 trajectories are analyzed only due to algorithmic complexity. 

Experiments on both synthetic and real data were performed to validate our approach. Robust Preference Analysis (RPA)~\cite{MagriFusiello15} was used in Step~1. A new dataset was also created, consisting of five sequences of moving objects in an indoor environment, where SIFT keypoints \cite{Lowe04} were extracted and manually labelled to get ground-truth segmentation. Results are reported in Sec.~\ref{sec:experiments}, where it is shown that: our method is comparable or better than most traditional (track-based) solutions on Hopkins155 \cite{TronVidal07}; it outperforms the methods developed in \cite{JiSalzmanLi15,XuCheongAl18} on synthetic/real datasets with mismatches; it is very effective in reducing the errors in the pairwise segmentations; it can be profitably used to segment SIFT keypoints in a collection of images.

Our two-step formulation of motion segmentation is inspired by the success gained by \emph{synchronization} or \emph{averaging} methods (e.g.~\cite{Singer11,TorselloRodolaAl11,HartleyAftabAl11,ArrigoniFusielloAl15siims,ChatterjeeGovindu17}) that formulate other computer vision problems (e.g.~structure from motion and multi-view registration) in an analogous manner. For instance, multi-view registration -- where the task is to bring multiple scans into alignment -- can be addressed by first computing rigid transformations between each pair of scans in isolation, and then globally optimizing these transformations without considering points.
In particular, our method -- which computes the segmentation of one image at a time (as explained in Sec.~\ref{sec:our_method}) -- presents similarities with \cite{TorselloRodolaAl11,HartleyAftabAl11}, which estimate the transformation of one camera/scan at a time.

\subsection{Related Work}
\label{sec:relatedwork}
\noindent Motion segmentation lies at the intersection of several computer vision problems, including subspace separation, multiple model fitting and multibody structure from motion.

The goal of \emph{subspace separation} is to cluster high-dimensional data drawn from multiple low-dimensional subspaces. Existing solutions include Generalized Principal Component Analysis (GPCA) \cite{VidalMaAl05}, Local Subspace Affinity (LSA) \cite{YanPollefeys06}, Power Factorization (PF) \cite{VidalTronAl08}, Agglomerative Lossy Compression (ALC) \cite{RaoTronAl10}, Low-Rank Representation (LRR) \cite{LiuLinAl13}, Sparse Subspace Clustering (SSC) \cite{ElhamifarVidal13}, Structured Sparse Subspace Clustering (S$^3$C) \cite{LiVidal15}, and Robust Shape Interaction Matrix (RSIM) \cite{JiSalzmanLi15}. Motion segmentation can be cast as subspace separation since -- under the affine camera model -- the point trajectories lie in the union of $d$ subspaces in $\mathbb{R}^{2n}$ of dimension at most 4, where $d$ denotes the number of motions and $n$ denotes the number of images. Subspace separation techniques can also be used to solve motion segmentation in two images under the perspective camera model, since corresponding points undergoing the same motion -- after a proper rearrangement of coordinates -- belong to a subspace of $\mathbb{R}^9$ of dimension at most 8, as observed in \cite{LiGuoAl13}.

The goal of \emph{multiple model fitting} is to estimate multiple models (e.g.~geometric primitives) that fit data corrupted by outliers and noise, without knowing which model each point belongs to. Some methods follow a consensus-based approach, namely they focus on the estimation part of the problem, with the aim of finding models that describe as many points as possible. The Hough transform \cite{XuOjaAl90}, Sequential RANSAC \cite{VincentLaganiere01}, Multi-RANSAC \cite{ZulianiKenneyAl05} and Random Sample Coverage (RansaCov) \cite{MagriFusiello16} belong to this category.
Other techniques follow a preference-based approach, namely they concentrate on the segmentation side of the problem, from which model estimation follows.
Solutions of this type include Residual Histogram Analysis (RHA) \cite{ZhangKosecka06}, J-Linkage \cite{ToldoFusiello08}, Kernel Optimization \cite{ChinSuterAl10}, T-linkage \cite{MagriFusiello14}, Random
Cluster Model (RCM) \cite{PhamChinAl14} and Robust Preference Analysis (RPA) \cite{MagriFusiello15}. The problem of fitting multiple models can also be expressed in terms of energy minimization \cite{DelongGorelickAl12,DelongOsokinAl12}, as done by PEARL (Propose Expand and Re-estimate Labels) \cite{IsackBoykov12} and Multi-X \cite{BarathMatas18}.
Model fitting techniques can be exploited to solve motion segmentation under the affine camera model, by fitting multiple subspaces to feature trajectories in an image sequence, similarly to subspace separation methods. They can also be used to solve motion segmentation in two images under the perspective camera model, by fitting multiple fundamental matrices to corresponding points in an image pair.

The goal of \emph{multibody structure from motion} is to simultaneously estimate the motion between each object and the camera as well as the 3D structure of each object, given a set of images of a dynamic scene. This problem can be seen as the generalization of structure from motion \cite{OzyesilVoroninskiAl17} to the dynamic case, where motion segmentation has to be solved in addition to 3D reconstruction. Geometric solutions are available for two images \cite{VidalMaAl06} and three images \cite{VidalHartley08}. 
Other techniques follow a statistical approach \cite{Torr98,QianChellappaAl05,SchindlerSuterAl08,ThakoorGaoAl10,OzdenSchindlerAl10,SabzevariScaramuzza14}, whereas in \cite{Gear98,CosteiraKanade98,LiKallemAl07,ZappellaDelbueAl13} motion segmentation and structure from motion are combined. More details can be found in survey \cite{SaputraMarkhamAl18}.

Ad-hoc solutions to motion segmentation are also present in the literature \cite{LiGuoAl13,LaiWangAl17,XuCheongAl18}, which are not explicitly related to the aforementioned problems. The authors of \cite{LiGuoAl13} formulate a joint optimization problem which builds upon the SSC algorithm, where it is required that all image pairs share a common sparsity profile. In \cite{LaiWangAl17} an accumulated correlation matrix is built by sampling homographies over consecutive image pairs, 
and spectral clustering \cite{Von-Luxburg07} is applied to get the sought segmentation. Such approach is generalized in \cite{XuCheongAl18} where multiple models (affine, fundamental and homography) are combined to get an improved segmentation. Different approaches are analyzed to reach such task, namely Kernel Addition (KerAdd) \cite{CortesMohriAl09}, Co-Regularization (Coreg) \cite{KumarRaiAl11} and Subset Constrained Clustering (Subset) \cite{WangQianAl14}. Motion segmentation is also addressed in \cite{SabzevariScaramuzza16,RubinoDelbueAl18}, where existing algorithms are customized for specific scenarios and acquisition platforms.

\section{Problem Formulation}
\label{sec:formulation}
\begin{figure}[htbp] 
  \centering
\includegraphics[width=0.23\linewidth]{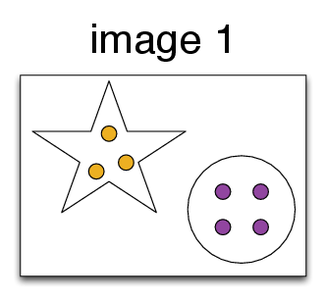} 
\includegraphics[width=0.23\linewidth]{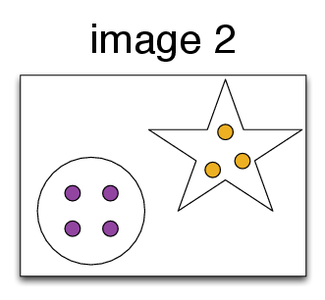} 
\includegraphics[width=0.23\linewidth]{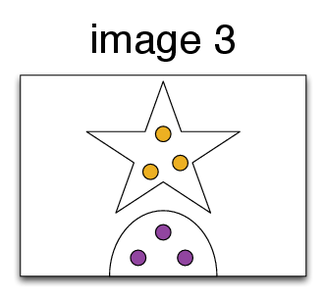}
\includegraphics[width=0.23\linewidth]{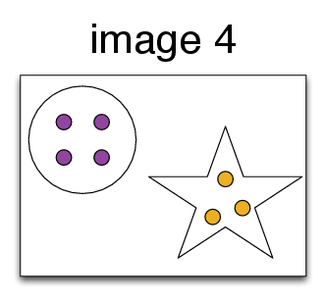} 
\caption{
A set of points is detected in multiple images and the goal is to assign them a label (purple or yellow) based on the moving object (star or circle) they belong to.
}
\label{fig:segmented_images}
\end{figure}

\noindent  Let $n$ denote the number of images and let $d$ denote the number of motions. Suppose that a number $p_i$ of points are found in image $i$ using a feature extraction algorithm, so that the total amount of points over all the images is given by $p=\sum_{i=1}^n p_i$. Let $\mathbf{s}_i \in \{0,1,\dots,d \}^{p_i}$ denote the labels of points in image $i$, which identify the membership to a specific motion. The meaning of the zero label, which essentially represents outliers, will be clarified in Sec.~\ref{sec:outliers}. The vector $\mathbf{s}_i$ is referred to as the \emph{total segmentation} 
of image~$i$, since it represents labels of points considering a \emph{global} numbering of motions. 
The goal here is to estimate $\mathbf{s}_i$ for $i=1,\dots,n$, as shown in Fig.~\ref{fig:segmented_images}. In other words, we aim at classifying image points as opposed to existing methods which segment tracks.
In order to accomplish such a task, we assume that points have been matched in image pairs and that segmentation between pairs of images is available. Note that the knowledge of matches, which involve two images at a time, is a weaker assumption than the presence of tracks, which involve all the images simultaneously. 

\begin{figure}[htbp] 
  \centering
\includegraphics[width=0.48\linewidth]{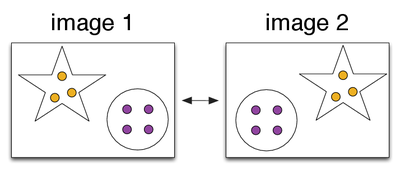} 
\includegraphics[width=0.48\linewidth]{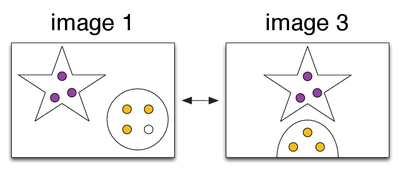} 
\includegraphics[width=0.48\linewidth]{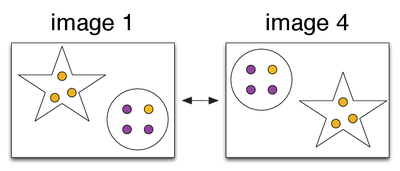} 
\includegraphics[width=0.48\linewidth]{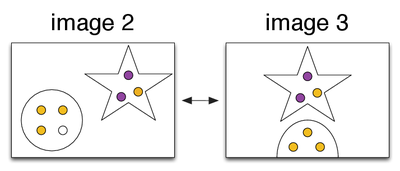} 
\includegraphics[width=0.48\linewidth]{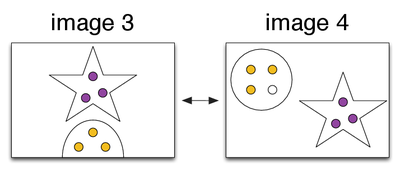} 
\includegraphics[width=0.48\linewidth]{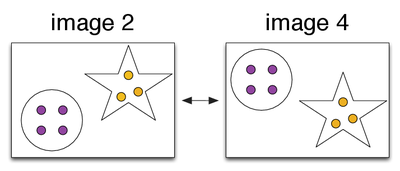} 
\caption{
Motion segmentation is performed on 
image pairs (with possible errors). The same motion (star or circle) may be given a different label (purple or yellow) in different pairs. 
}
\label{fig:pairwise_segmentation}
\end{figure}

Let $\mathbf{s}_{ij}\in \{0,1,\dots,d \}^{m_{ij}}$ denote the labels of corresponding points in images $i$ and $j$, where the zero label corresponds to outliers and let $m_{ij} \le \min \{ p_i,p_j \}$ denote the number of matches of the pair $(i,j)$. Vector $\mathbf{s}_{ij}$ is referred to as the \emph{partial segmentation} of the pair $(i,j)$, since it represents labels of corresponding points considering a \emph{local} numbering of motions, as shown in Fig.~\ref{fig:pairwise_segmentation}. 
Observe that each $\mathbf{s}_{ij}$ may contain some errors, which can be caused either by mismatches or by failure of the algorithm used for pairwise segmentation, and some image points may not have a label assigned in some pairs due to missing correspondences.

Thus we have to face the problem of how to assign a unique/global label to all image points such that the constraints coming from pairwise segmentation are best satisfied. In other words, the segmentation task can be reduced to the problem of estimating the total segmentations $\mathbf{s}_1, \dots, \mathbf{s}_n$ starting from the knowledge of partial segmentations $\mathbf{s}_{ij}$ with $i,j=1, \dots, n$. It is worth noting that in this way the actual coordinates of image points are not used anymore after pairwise segmentation, since only labels matter for the final segmentation. Observe also that this general formulation does not assume any particular camera model or scene geometry.
%
\section{Proposed Method}\label{sec:our_method}
\begin{figure}[htbp] 
  \centering
\includegraphics[width=1\linewidth]{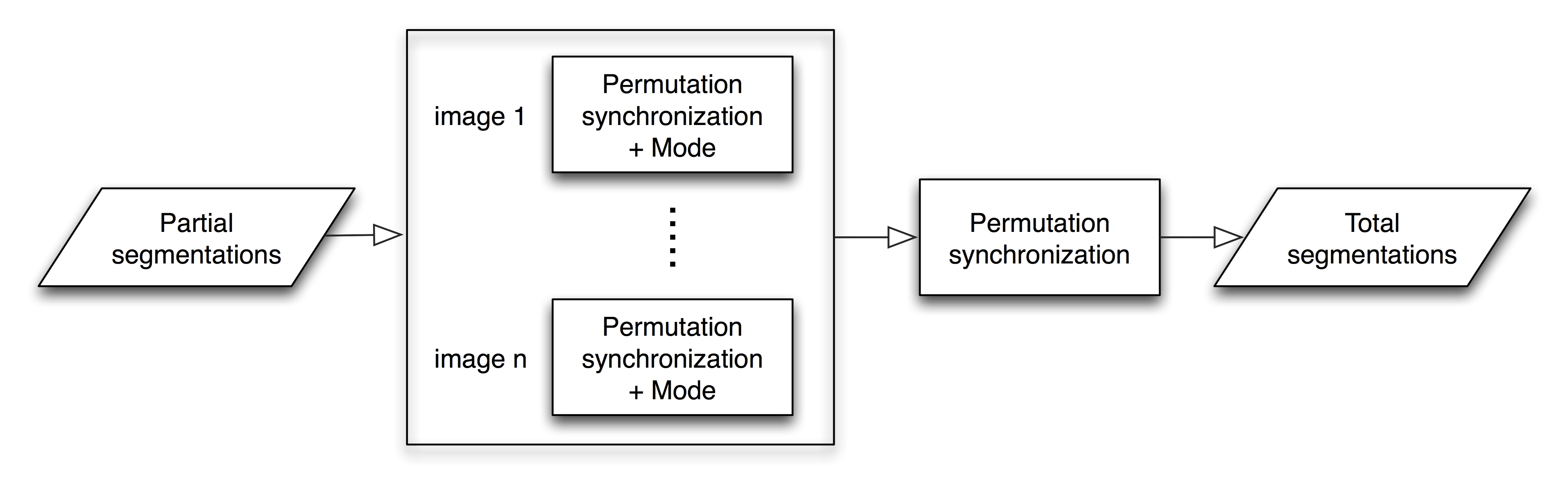}
\caption{
Outline of the proposed approach.
}
\label{fig:outline}
\end{figure}

\noindent Our method (sketched in Fig.~\ref{fig:outline}) takes as input the results from pairwise segmentation. It first computes the total segmentation of each image individually and then updates all these estimates in order to have a single/global numbering of motions. 
\subsection{Segmenting a single image}
\begin{figure}[htbp] 
  \centering
\includegraphics[width=0.23\linewidth]{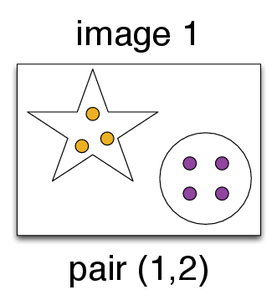} 
\includegraphics[width=0.23\linewidth]{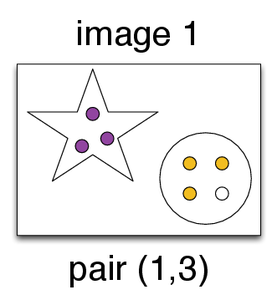} 
\includegraphics[width=0.23\linewidth]{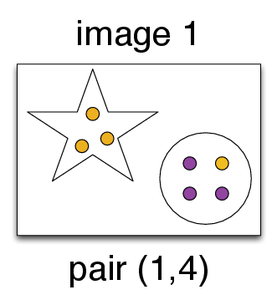} 
\caption{A possible solution for the total segmentation of image 1 is given by each partial segmentation where image 1 is involved. The same motion (star or circle) may be given a different label (purple or yellow) in different pairs.
}
\label{fig:multiple_segmentations}
\end{figure}

\noindent  The key observation is that each partial segmentation $\mathbf{s}_{ij} \in \{0,1,\dots,d \}^{m_{ij}}$ gives rise to two vectors 
\begin{gather}
\widehat{\mathbf{s}}_i^j \in \{ \text{NaN},0,1,\dots,d \}^{p_i} \\ 
\widehat{\mathbf{s}}_j^i \in \{ \text{NaN},0,1,\dots,d \}^{p_j}
\end{gather}
which contain labels of matching points in images $i$ and $j$, respectively, where NaN accounts for missing correspondences. This implies that, if we fix one image (e.g.~image $i$), then several estimates are available for its total segmentation, which define a set $ \mathcal{B}_i $
\begin{equation}
\mathcal{B}_i = \{ \widehat{\mathbf{s}}_i^k \text{ s.t. } k=1, \dots, n , \ k \neq i \}.
\label{eq:several_estimates}
\end{equation}
However, these estimates are not absolute since they may differ by a permutation of the labels associated with each motion, as shown in Fig.~\ref{fig:multiple_segmentations}.

In order to resolve such ambiguity, we consider a graph where each node is an element in $ \mathcal{B}_i $ (i.e.\ a partial segmentation involving image $i$) and the edge between nodes $h$ and $k$ is associated with a permutation $P_{hk}$ of labels that best maps $\widehat{\mathbf{s}}_i^k$ (i.e.\ labels of image $i$ in the pair $(i,k)$) into $\widehat{\mathbf{s}}_i^h$ (i.e.\ labels of image $i$ in the pair $(i,h)$). Computing such permutation is a \emph{linear assignment problem}, which can be solved using the Hungarian algorithm \cite{Kuhn55}. The task here is to compute a permutation $P_k$ for each node that reveals the true numbering of motions. It can be seen that this can be expressed as a \emph{permutation synchronization}, that is the problem of estimating $P_k$ for $k = 1, \dots , n $ ($ k \neq i $) such that $P_{hk} = P_h P_k^{-1}$, which can be solved via eigenvalue decomposition \cite{PachauriKondorAl13}.
%
%
\begin{figure}[htbp] 
  \centering
\includegraphics[width=0.23\linewidth]{Figure/image1_pair_12} 
\includegraphics[width=0.23\linewidth]{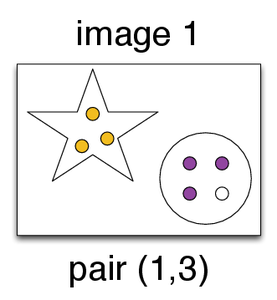} 
\includegraphics[width=0.23\linewidth]{Figure/image1_pair_14} 
\caption{
After solving a permutation synchronization problem, several estimates for the total segmentation of image 1 are available, where the same motion (star or circle) has the same label (purple or yellow) in different pairs.
}
\label{fig:multiple_segmentations_aligned}
\end{figure}

After this step, the set in Eq.~\eqref{eq:several_estimates} contains several estimates of $\mathbf{s}_i$ with respect to a single numbering of motions, as shown in Fig.~\ref{fig:multiple_segmentations_aligned}.
Thus a scheme that assigns a unique label to each point in image $i$ is required, which can be regarded as the best over the set $ \mathcal{B}_i $.
A reasonable approach consists in labelling each point with the most frequent label (i.e.~the \emph{mode}) among all the available measures.
In other words, the label of point $r$ is given by
\begin{equation}
    \mathbf{s}_i(r) = \text{mode } \{ \widehat{\mathbf{s}}_i^k (r) \text{ s.t. } \widehat{\mathbf{s}}_i^k \in \mathcal{B}_i, \ \widehat{\mathbf{s}}_i^k (r) \neq \text{NaN} \}
    \label{eq:mode}
\end{equation}
where only labels of actual correspondences are considered, with $r=1, \dots, p_i$.  
%
As long as the algorithm used for pairwise segmentation correctly classifies all the points in most pairs, this procedure works well, as confirmed by experiments in Sec.~\ref{sec:experiments}.
%
\subsection{Segmenting multiple images}
\begin{figure}[htbp] 
  \centering
\includegraphics[width=0.23\linewidth]{Figure/image1} 
\includegraphics[width=0.23\linewidth]{Figure/image2} 
\includegraphics[width=0.23\linewidth]{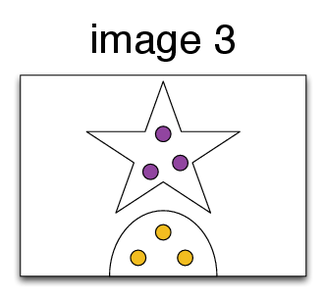} 
\includegraphics[width=0.23\linewidth]{Figure/image4} 
\caption{
Motion segmentation is performed on each image individually. The same motion (star or circle) may be given a different label (purple or yellow) in different images.
}
\label{fig:image_final}
\end{figure}

\noindent The above procedure is applied to all the images in order to estimate the sought total segmentations $\mathbf{s}_1, \mathbf{s}_2 \dots, \mathbf{s}_n$. Such estimates, however, are not absolute since each image has been treated independently from the others, and hence results may differ by a permutation of the labels associated with each motion, as shown in Fig.~\ref{fig:image_final}.

In order to address this issue, we consider a graph where each node corresponds to an image and the edge between images $i$ and $j$ is associated with a permutation $P_{ij}$ that best maps $\mathbf{s}_j$ into $\mathbf{s}_i$. In order to compute such permutation, we ground on pairwise segmentation, since labels of the same points are required: in order to map $\mathbf{s}_j$ (labels of image $j$) into $\mathbf{s}_i$ (labels of image $i$), we first map $\widehat{\mathbf{s}}_i^j$ (labels of image $i$ in the pair $(i,j)$) into $\mathbf{s}_i$, and then we map $\mathbf{s}_j$ into $\widehat{\mathbf{s}}_j^i$ (labels of image $j$ in the pair $(i,j)$). These are linear assignment problems \cite{Kuhn55}. Thus the task is to compute a permutation $P_i$ for each image that reveals the true numbering of motions such that $ P_{ij} = P_i P_j^{-1} $, which can be viewed as a permutation synchronization \cite{PachauriKondorAl13}. Hence all the total segmentations are expressed with respect to the same numbering of motions, as in Fig.~\ref{fig:segmented_images}.
%
%
%
\begin{table*}[htbp]
\centering
\caption{Average misclassification error [$\%$] for several methods on the Hopkins155 benchmark \cite{TronVidal07}. Results are copied from \cite{XuCheongAl18}.
 \label{tab_hopkins}}
 \resizebox{1\textwidth}{!}{
\begin{tabular}{ lccccccccccccccccc } 
\hline\noalign{\smallskip}
& LSA	 \cite{YanPollefeys06}& GPCA \cite{VidalMaAl05} & ALC \cite{RaoTronAl10} & SSC \cite{ElhamifarVidal13} & TPV \cite{LiGuoAl13} & LRR \cite{LiuLinAl13} & T-Linkage \cite{MagriFusiello14} & S$^3$C \cite{LiVidal15} & RSIM \cite{JiSalzmanLi15} & MSSC \cite{LaiWangAl17} & KerAdd \cite{XuCheongAl18} & Coreg \cite{XuCheongAl18}  & Subset \cite{XuCheongAl18} & Baseline & \ourmethod \\
\noalign{\smallskip}
\hline
\noalign{\smallskip} 
 2 Motions & 4.23 & 4.59 & 2.40 & 1.52 & 1.57& 1.33& 0.86& 1.94& 0.78& 0.54& 0.27& 0.37& \bf 0.23 & 2.26 & 1.00 \\
 3 Motions & 7.02 & 28.66 &6.69 &4.40 &4.98 &4.98 &5.78 &4.92 &1.77 &1.84 &0.66 &0.75 & \bf 0.58  & 9.04 & 2.67 \\
 All & 4.86 & 10.02 & 3.56 & 2.18 & 2.34 & 1.59 & 1.97 & 2.61 & 1.01 & 0.83 & 0.36 & 0.46 & \bf 0.31 & 3.79 & 1.37 \\
\noalign{\smallskip}
\hline
\end{tabular}
}
\end{table*}
\begin{table*}[htbp]
\centering
\caption{Average and median misclassification error [$\%$] for several methods on the Hopkins12 benchmark \cite{VidalTronAl08}. 
Results for different variants of ALC and SSC are taken from \cite{JiSalzmanLi15} whereas results for the remaining methods are copied from the respective papers.
 \label{tab_hopkins12}}
 \resizebox{1\textwidth}{!}{
\begin{tabular}{ lccccccccccccccccc } 
\hline\noalign{\smallskip}
& PF \cite{VidalTronAl08} & PF+ALC \cite{RaoTronAl10} & RPCA+ALC \cite{RaoTronAl10} & $\ell_1$+ALC \cite{RaoTronAl10}  & SSC-R \cite{ElhamifarVidal13} & SSC-O \cite{ElhamifarVidal13} & RSIM \cite{JiSalzmanLi15} & KerAdd \cite{XuCheongAl18} & Coreg \cite{XuCheongAl18}  & Subset \cite{XuCheongAl18}  & Baseline &\ourmethod \\
\noalign{\smallskip}
\hline
\noalign{\smallskip} 
 Mean & 14.94 & 10.81 & 13.78 & 1.28 & 3.82 & 8.78 & 0.61 &  0.11   & \bf0.06 & \bf 0.06 & 7.45 & 4.33 \\
 Median & 9.31 & 7.85 & 8.27 & 1.07 & 0.31 & 4.80 & 0.61 &\bf 0.00 & \bf0.00 & \bf 0.00  & 2.16 & 0.38 \\
\noalign{\smallskip}
\hline
\end{tabular}
}
\end{table*}
\subsection{Dealing with outliers}
\label{sec:outliers}
\noindent When doing pairwise segmentation, it is expected that mismatched points are classified as outlier (zero label). When dealing with total segmentation, instead, the situation is different: in principle, there exists no outlier since each image point actually belongs to a motion. However, in the presence of high corruption in the input matches, one may not be able to assign a valid label to all image points. Indeed, it may happen that a point is mismatched (and hence assigned the zero label) in all the pairs, so that there is no valid information to classify it. Such points are expected to have zero label in the absolute segmentation. However, since they are not actual outliers, we will refer to them as ``unclassified'' or ``unknown'' in the experiments.
%

In order to deal with those points, a reasonable approach is to ignore the labels which are set to zero by pairwise segmentation and compute the mode over the remaining measures, i.e.~substitute them with NaN before using Eq.~\eqref{eq:mode}. In this way all the image points are assigned a valid label (except those which are deemed as outlier in all the pairs), meaning that this approach tends to classify a high amount of points even in the presence of mismatches.
%
\section{Experiments}
\label{sec:experiments}
\noindent  In order to evaluate the performance of our approach -- named \ourmethod\footnote{The Matlab code will be made available on the web.} -- we ran experiments on both synthetic data and real images, in addition to the real data Hopkins155 \cite{TronVidal07} and Hopkins12 \cite{VidalTronAl08}. For pairwise segmentations -- which constitute the input to our method -- we fitted multiple fundamental matrices to correspondences in each image pair using RPA \cite{MagriFusiello15} (code available online\footnote{\scriptsize\url{http://www.diegm.uniud.it/fusiello/demo/rpa/}}).
Default values specified in the original paper were used for the algorithmic parameters in all the experiments.

Note that there are no direct competitors to our method, since the task of segmentation from pairwise matches has not been addressed so far. For this reason, we will focus on the comparison with a trivial solution (named the ``baseline'') which takes the \emph{same} input as our approach (i.e.~the results from pairwise segmentation) and it is constructed as follows: first, a maximum-weight spanning tree is computed, where each node in the graph is an image and edges are weighted with the number of inliers; then, the results from pairwise segmentation are used to segment each image along the tree, where the global numbering of motions is fixed at the root and propagated to the leaves.

Similarly to most works in motion segmentation literature, we assume that the number of motions is known in advance and give this value as input to all the analysed techniques.

\subsection{Hopkins Datasets}
\label{sec:hopkins}
\noindent  The Hopkins155 benchmark \cite{TronVidal07} contains 155 sequences of indoor and outdoor scenes with two or three motions, which are categorized into checkerboard, traffic and articulated/nonrigid sequences, and the Hopkins12 dataset \cite{VidalTronAl08} provides 12 additional sequences with missing data. We emphasize that these datasets provide (cleaned) tracks over multiple images, so they are not suitable for the task addressed in this paper, which is segmentation from raw pairwise matches. However, we report results on these sequences since they are widely used in segmentation literature. 

In order to make a meaningful comparison with the state of the art, a scheme that assigns a unique label to each track is required, starting from labels of image points. To accomplish such a task, we use the same criterion as the one developed in Sec.~\ref{sec:our_method} to label each image point given multiple measures derived from pairwise segmentation. We assign to each track the mode of the labels of points belonging to the track, and the same procedure is applied to the baseline.
Performance is measured in terms of \emph{misclassification error}, that is the percentage of misclassified tracks, as it is customary in motion segmentation literature. Tracks labelled as zero (if any) were counted as errors, since we know that outliers are not present in these datasets.
\begin{figure}[htbp] 
  \centering
   \subfloat[Hopkins12]{
\includegraphics[width=0.4\linewidth]{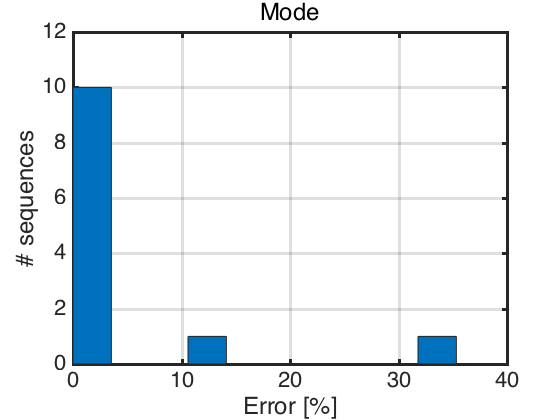}
\includegraphics[width=0.4\linewidth]{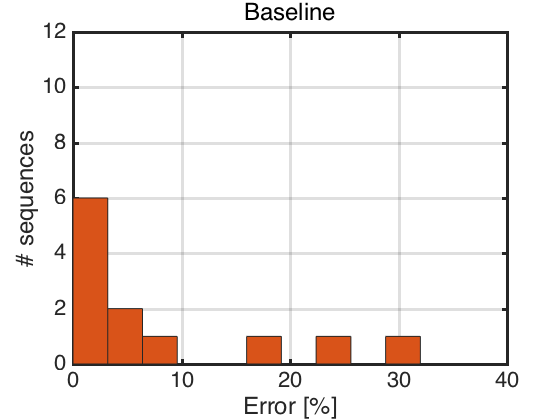}
} \\
   \subfloat[Hopkins155]{
\includegraphics[width=0.4\linewidth]{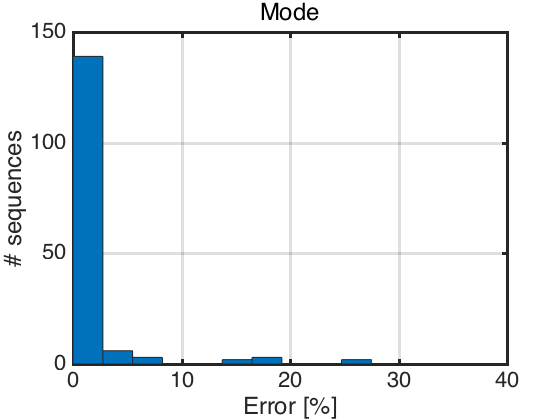}
\includegraphics[width=0.4\linewidth]{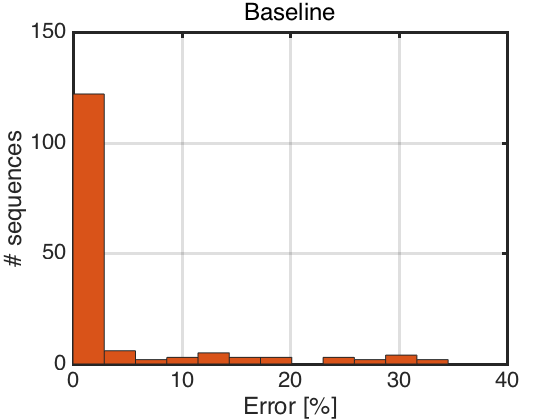}
}
\caption{
Histograms of misclassification errors achieved by \ourmethod and the baseline on the Hopkins155 \cite{TronVidal07} and Hopkins12 \cite{VidalTronAl08} datasets. The horizontal axis corresponds to a possible misclassification error in an individual sequence, and the vertical axis corresponds to the number of sequences where a given error is reached.
}
\label{fig:Hopkins}
\end{figure}

Results are reported in Tab.~\ref{tab_hopkins} and Tab.~\ref{tab_hopkins12} where \ourmethod is compared to several motion segmentation algorithms.  
Our approach clearly outperforms the baseline and it performs comparably or better than most of the state-of-the-art techniques, with a mean error of $1.37\%$ over all the sequences in Hopkins155 and a median error of $0.38\%$ over all the sequences in Hopkins12. 
In particular, it is noticeable that our method achieves (nearly) zero error in 139 out of 155 sequences in Hopkins155 and in 10 out of 12 sequences in Hopkins12, as shown in Fig.~\ref{fig:Hopkins}. After inspecting the solution, it was found that the remaining sequences correspond to situations where the algorithm used for pairwise segmentation (RPA) performed bad in most image pairs.

The fact that our method is not the best is not surprising since we are making much weaker assumptions (matches between image pairs instead of tracks over multiple images), i.e., we are addressing a more difficult task. Nevertheless, our method achieves good performances. In general, there is no reason to use our approach when tracks are available and one out of the best traditional methods (e.g.\cite{JiSalzmanLi15,LaiWangAl17,XuCheongAl18}) can be used. \emph{Our method is designed for the scenario where pairwise matches are available only}. The next sections demonstrate the benefits of our approach for this specific task.
%
\subsection{Simulated Data}
\label{sec:synthetic_experiments}
\noindent  We considered the \emph{cars1} dataset from the traffic sequences in Hopkins155, where $d=2$, $n=20$ and $p=6140$. 
Noise-free pairwise matches were obtained from the available tracks and synthetic errors were added to these correspondences in order to produce mismatches. More precisely, in each image pair a fraction of the correspondences -- which ranged from 0 to 0.8 in our experiments -- was randomly switched. This scenario resembles unordered image collections (e.g. in multibody structure from motion) where errors are ubiquitous among pairwise matches. For each configuration the test was repeated 10 times and average results were computed.

\begin{figure}[htbp] 
  \centering
\includegraphics[width=0.46\linewidth]{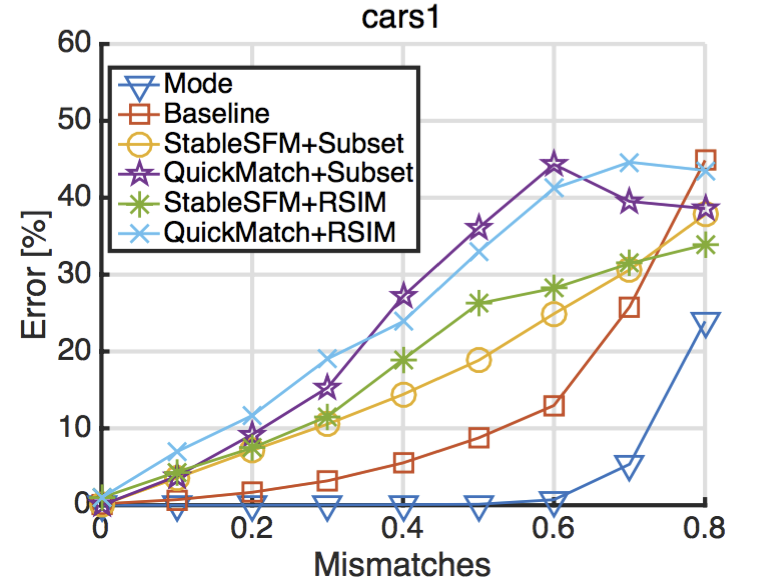} \quad 
\includegraphics[width=0.46\linewidth]{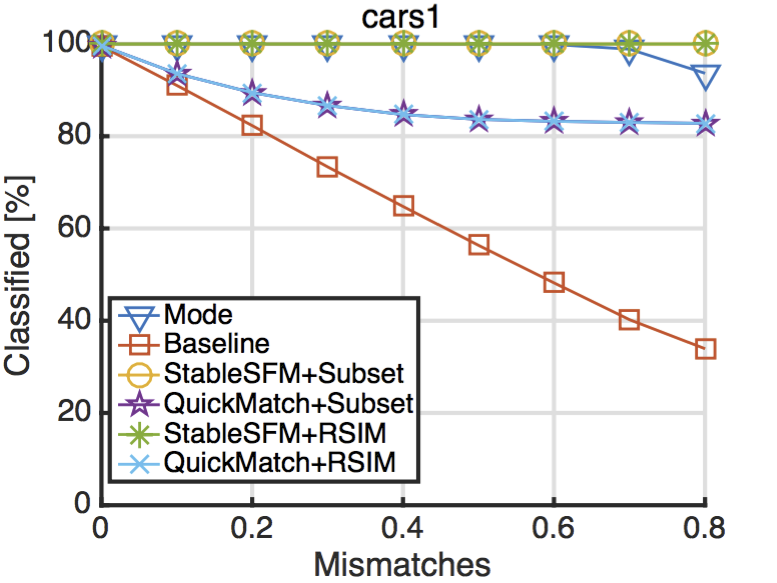} 
\caption{
Misclassification error [$\%$] and classified points [$\%$] versus fraction of mismatches for several methods on the \emph{cars1} sequence from Hopkins155 \cite{TronVidal07}.
}
\label{fig:missrate}
\end{figure}
\begin{figure}[htbp] 
  \centering
   \includegraphics[width=0.32\linewidth]{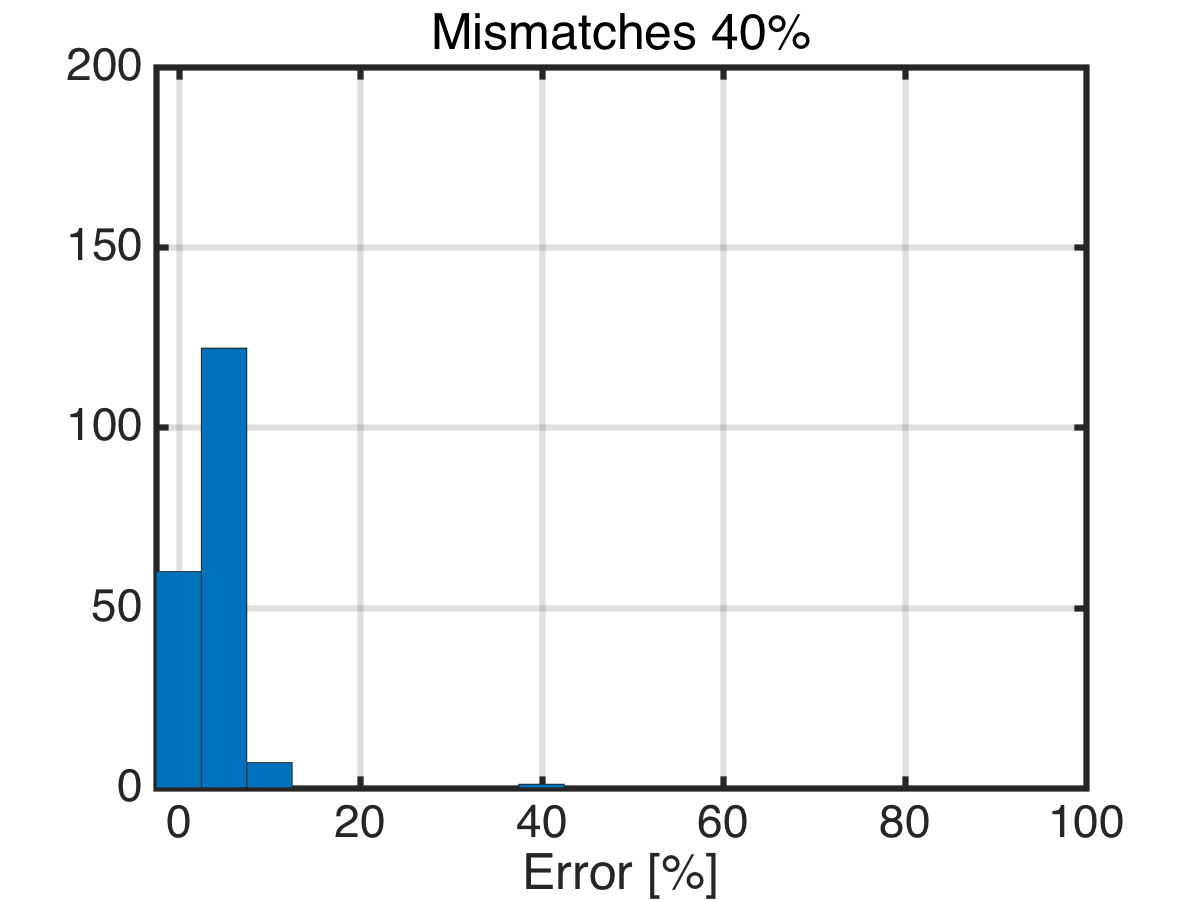}
   \includegraphics[width=0.32\linewidth]{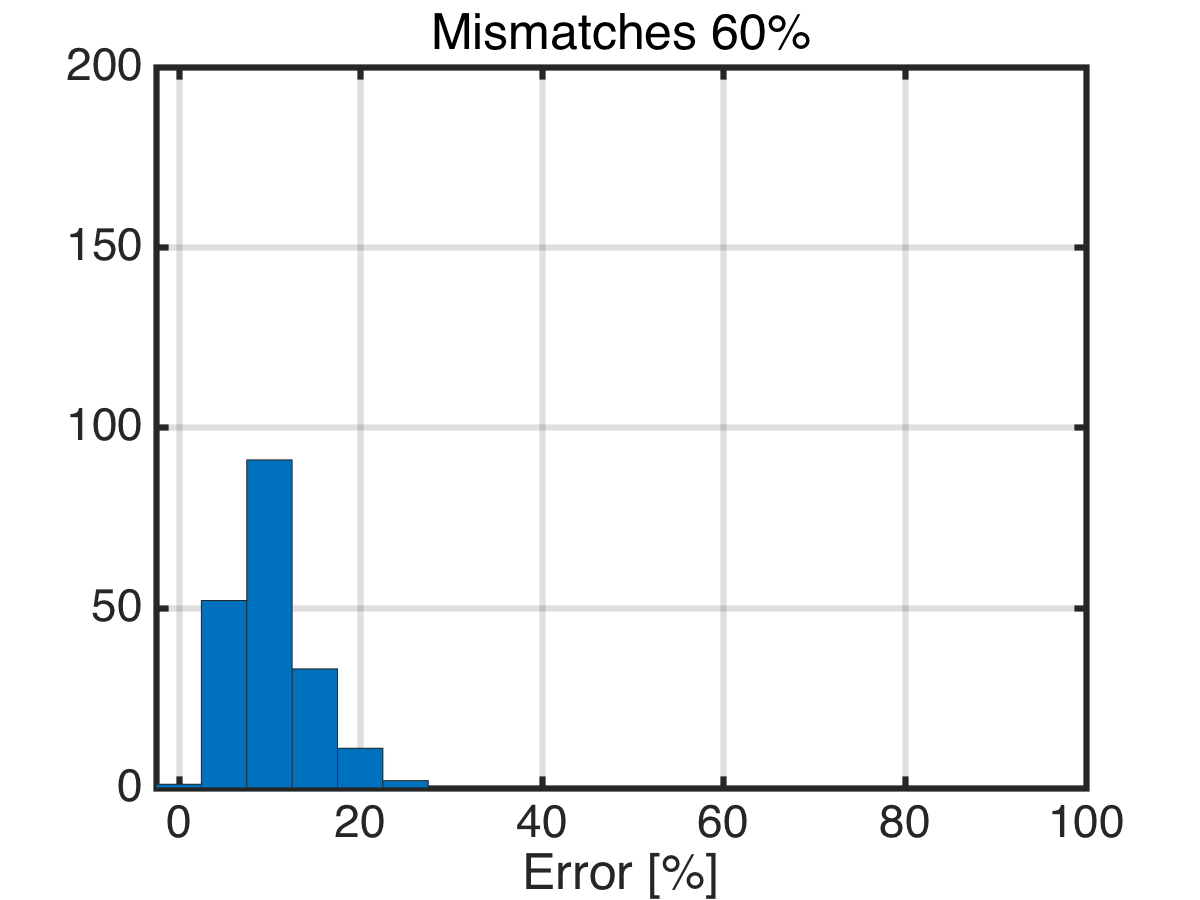}
   \includegraphics[width=0.32\linewidth]{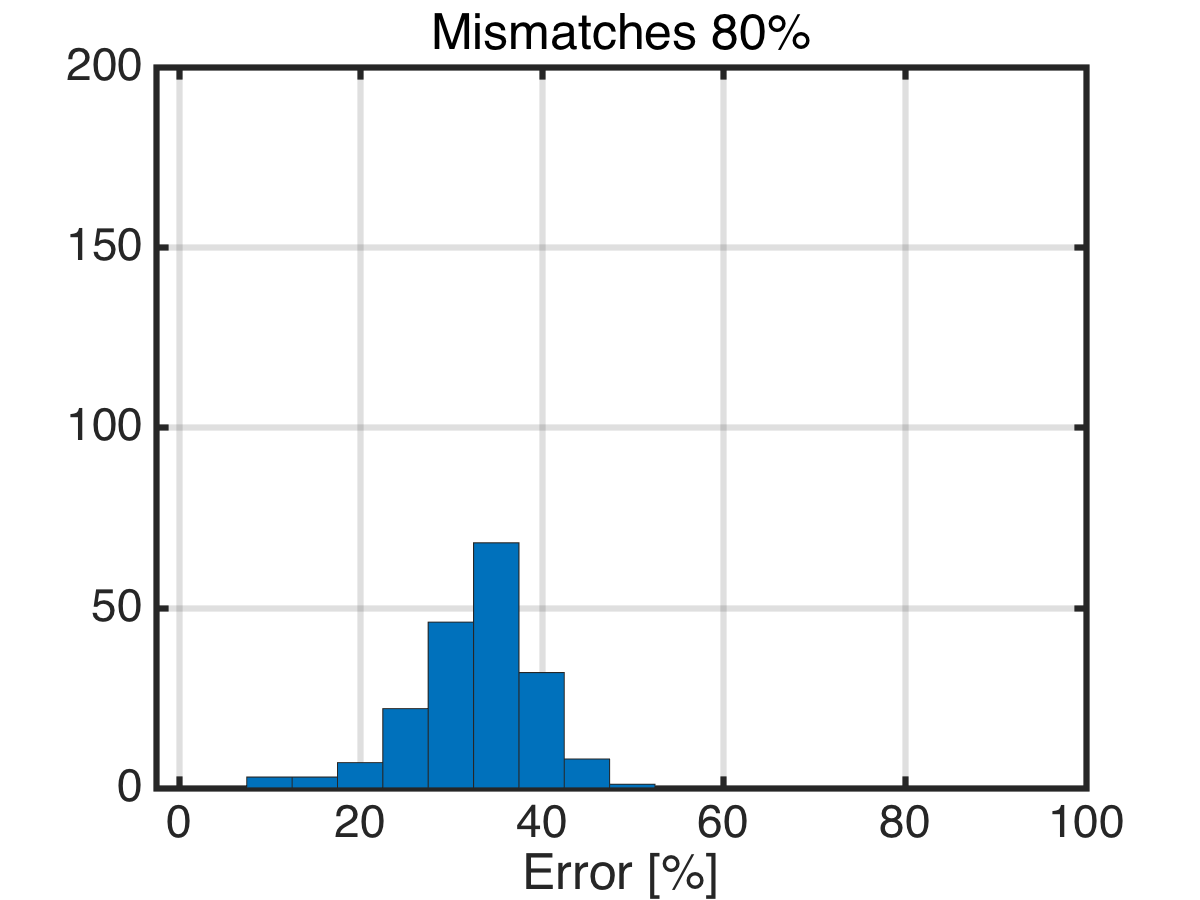}
\caption{
Histograms of misclassification error achieved by RPA~\cite{MagriFusiello15} on \emph{cars1} \cite{TronVidal07} for a single trial. The horizontal axis corresponds to the misclassification error in an individual image pair. The vertical axis corresponds to the number of pairs where a given error is obtained.
}
\label{fig:histograms}
\end{figure}

We compared \ourmethod with the baseline, which -- as our method -- takes as input the results from pairwise segmentation. We also included in the comparison two traditional methods which require tracks over multiple images as input, namely RSIM\footnote{\scriptsize \url{https://github.com/panji1990/Robust-shape-interaction-matrix}} \cite{JiSalzmanLi15} and Subset\footnote{\scriptsize \url{https://alex-xun-xu.github.io/ProjectPage/CVPR_18/}} \cite{XuCheongAl18}, whose implementations are available online. The former provides a robust solution to subspace separation, whereas the latter can be regarded as the current state of the art in motion segmentation with mean error of $0.31\%$ on the Hopkins155 benchmark (see Tab.~\ref{tab_hopkins}). 
We used two different techniques for computing tracks from pairwise matches, namely StableSfM\footnote{\scriptsize \url{http://www.maths.lth.se/matematiklth/personal/calle/sys_paper/sys_paper.html}} \cite{OlssonEnqvist11} and QuichMatch\footnote{\scriptsize
\url{https://bitbucket.org/tronroberto/quickshiftmatching}} \cite{TronZhouAl17}.

Performance was measured in terms of misclassification errors, which is defined here as the percentage of misclassified points over the total amount of classified image points.  In other words, unlike in Sec.~\ref{sec:hopkins}, segmentation results were evaluated considering only points with a nonzero label (i.e.~points with zero label do not contribute to the error). Indeed, due to the presence of mismatches, one may not expect to give a valid label to all the image points, as observed in Sec.~\ref{sec:outliers}. We also computed the percentage of points classified by each method.

\begin{figure*}[htbp] 
\centering 
\subfloat[Mismatches $0\%$ \label{fig:bars0}]{
\includegraphics[width=0.7\linewidth]{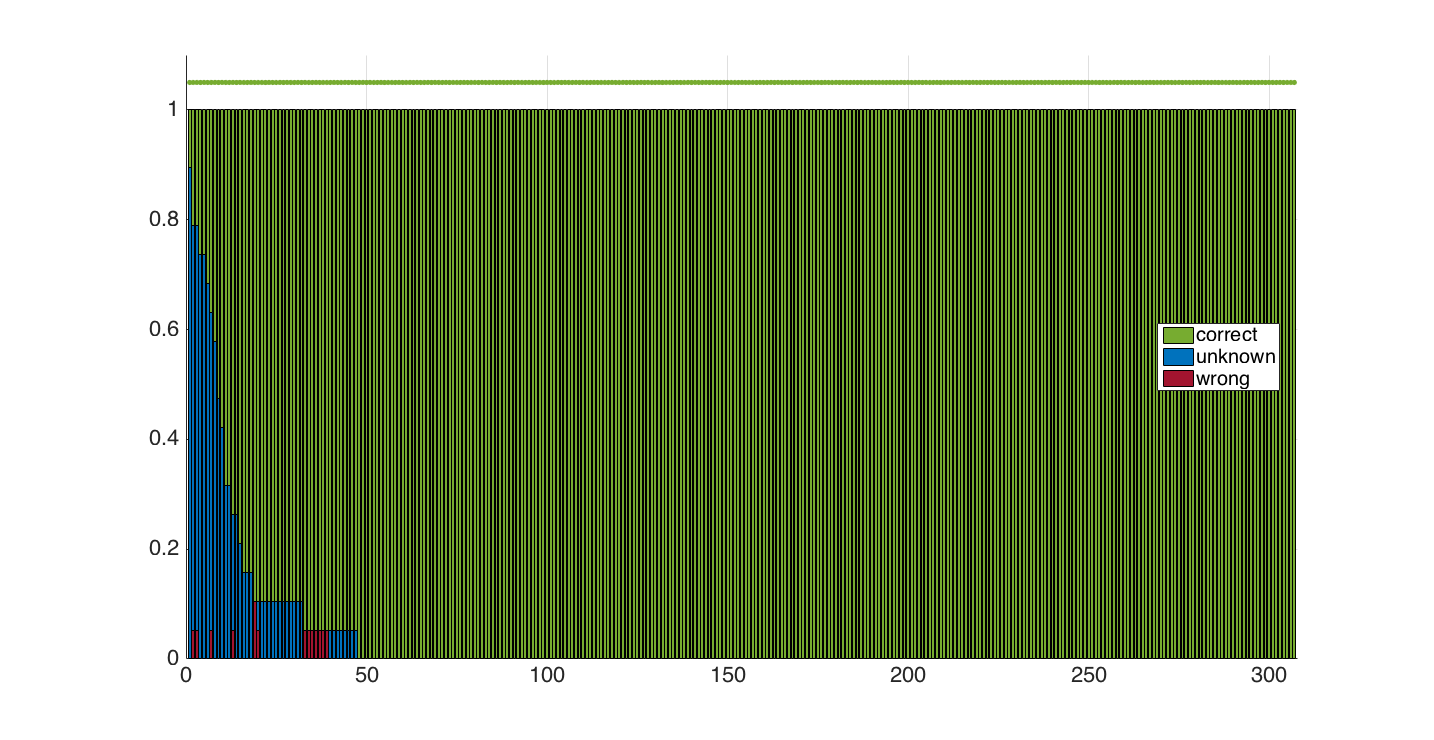}
} \\
\subfloat[Mismatches $40\%$]{
\includegraphics[width=0.7\linewidth]{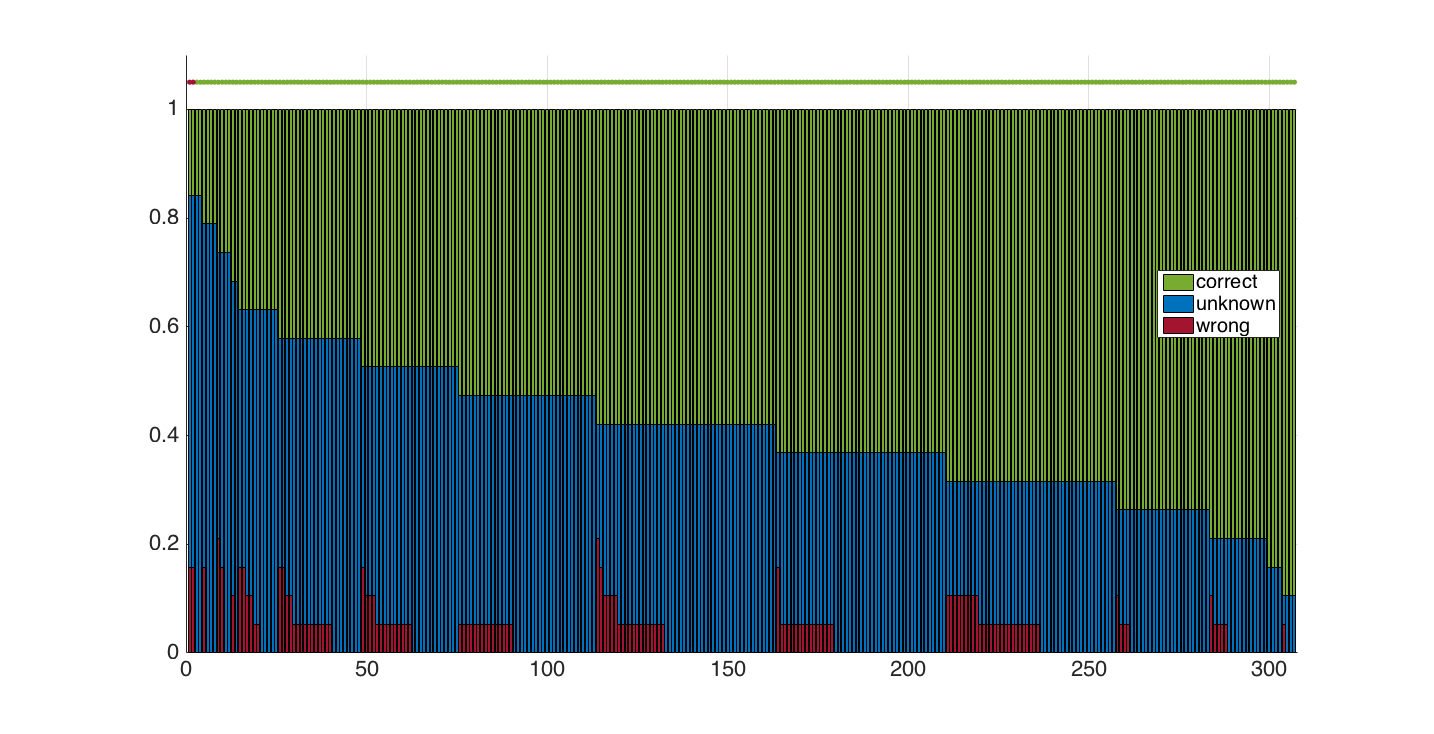}
} \\
\subfloat[Mismatches $80\%$]{
\includegraphics[width=0.7\linewidth]{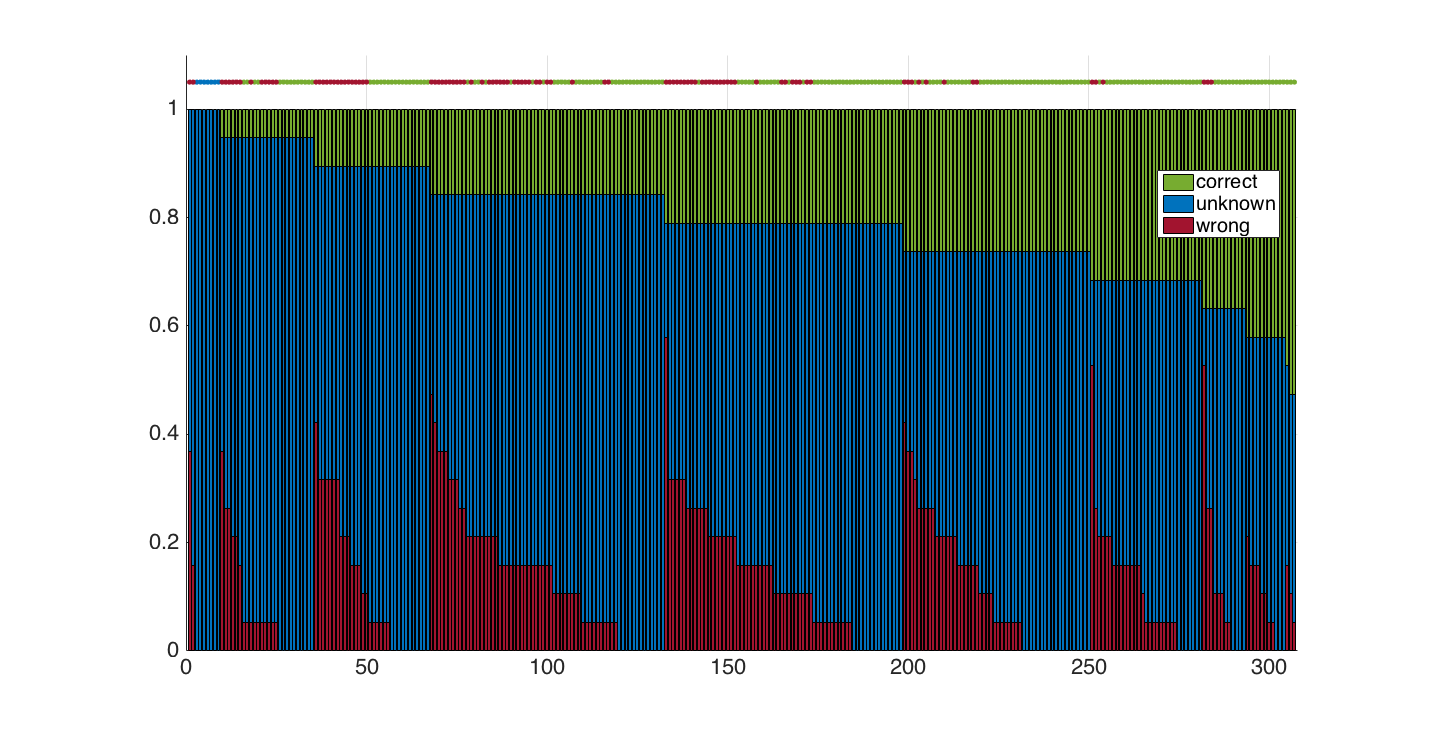}
}
\caption{
The horizontal axis indexes points in a sample image from \emph{cars1} \cite{TronVidal07} and a threee-color bar is shown for each point.
Bars are divided into three parts which sum to one. The green, red, and blue parts represent fractions of image pairs where the point is correctly classified, misclassified, and labeled as outlier, respectively, by RPA \cite{MagriFusiello15}. For better visualization, points are sorted increasingly by the height of green bars.
A dot is plotted over each bar to show whether the point is classified by our method correctly (green), misclassified (red) or labelled as unknown (blue).
}
\label{fig:bars}
\end{figure*}

Results are reported in Fig.~\ref{fig:missrate}, which clearly shows the robustness to mismatches gained by our approach: it is remarkable that the error remains constant (around $0\%$) with up to $60\%$ of mismatches. \ourmethod is significantly better than the baseline both in terms of misclassification error and percentage of classified points. The former exploits redundant measures in order to produce the final segmentation, whereas the latter uses results from a maximum-weight spanning tree only.

Concerning traditional methods, it was found by inspecting the solution that Subset and RSIM actually segment all the tracks, and unclassified data correspond to image points that were not included in any track by the algorithm used for computing tracks. Such techniques achieve a low misclassification error only when mismatches are below $10\%$ and performances degrade with increasing ratio of mismatches. Indeed, wrong correspondences propagate into the tracks making traditional motion segmentation really hard to solve. Notice that a track can even contain points of different motions, in which case errors in the output segmentation appear by assigning a unique label to the entire track. This clearly motivates the need of our method for segmentation from raw pairwise matches. 

In order to give a full picture on the performance of our approach, we report in Fig.~\ref{fig:histograms} the histograms of misclassification error achieved by RPA over all the image pairs, which gives an idea about how hard it is to solve the motion segmentation \emph{given} results of pairwise segmentation.
Indeed, RPA may fail to detect errors in the input matches and it may not correctly segment some points since it lacks theoretical guarantees, thus producing errors in the individual pairwise segmentations.
As expected, the histograms shift to the right as the percentage of input mismatches increases. 
Let us consider the central histogram, which corresponds to 60$\%$ of mismatches: it is worth noting that, despite individual pairwise segmentations are noisy, our method achieves zero error, as shown in Fig.~\ref{fig:missrate}. In other words, \ourmethod is able to successfully solve motion segmentation while reducing errors in the pairwise segmentations, thanks to the fact that it exploits redundant measures in a principled manner.

We now illustrate what happens to individual points when running our method. 
%
%
Figure \ref{fig:bars} reports coloured bars representing the amount of errors for each point in a sample image. 
As the percentage of mismatches increases, motion segmentation gets harder to solve, since the green area reduces whereas the blue and red ones enlarge.
Note that RPA \cite{MagriFusiello15} produces errors even in the absence of wrong correspondences, as can be appreciated in Fig.~\ref{fig:bars0}.
Our method classifies all the data except for a few cases where the blue bars are equal to 1, meaning that the point is labelled as outlier by RPA in all the pairs.
Among the classified points, \ourmethod provides a correct segmentation as long as the green bars are sufficiently high.

\subsection{Real Data}
\label{sec:real_experiments}
\noindent In order to evaluate the performance of our approach on real data, we considered both indoor and outdoor images. SIFT keypoints \cite{Lowe04} were extracted in all the images and correspondences between image pairs were established using the nearest neighbor and ratio test as in \cite{Lowe04}, using the VLFeat library\footnote{\scriptsize\url{http://www.vlfeat.org/}}.
For each image pair $(i,j)$, we kept only those correspondences that were found both when matching image $i$ with $j$ and when matching image $j$ with $i$, and isolated features (i.e.~points that are not matched in any image) were removed. No further filtering was applied.
\subsubsection{Indoor scenes}
\label{sec:real_exp_indoor}
\begin{figure}[htbp] 
  \centering
  \resizebox{1\columnwidth}{!}{
  \subfloat[Flowers]{
  \includegraphics[width=0.23\linewidth]{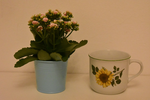}
  }
    \subfloat[Pencils]{
  \includegraphics[width=0.23\linewidth]{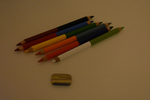}
  }
  \subfloat[Bag]{
  \includegraphics[width=0.23\linewidth]{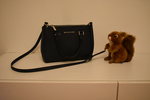}
  }
\subfloat[Bears]{
  \includegraphics[width=0.23\linewidth]{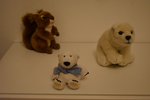}
  }
  \subfloat[Penguin]{
  \includegraphics[width=0.23\linewidth]{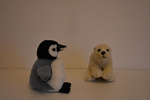}
  }
  }
\caption{
Sample images from our dataset.
}
\label{fig:our_dataset}
\end{figure}

\begin{figure}[htbp] 
  \centering
\includegraphics[width=0.6\linewidth]{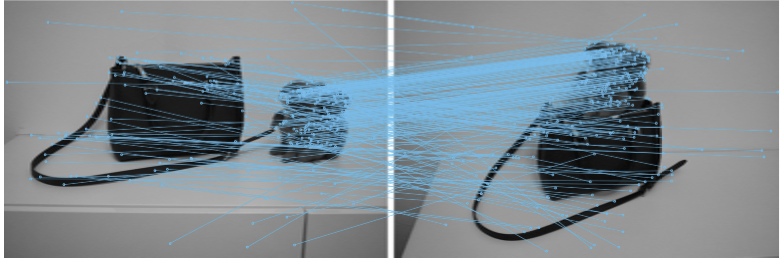}
\caption{
SIFT matches on an image pair from the \emph{Bag} sequence.
}
\label{fig:mismatches}
\end{figure}

\noindent Since there are no standard datasets for segmentation from pairwise matches, we created a small benchmark\footnote{The dataset will be made available on the web.} consisting of five image sequences. We considered indoor scenes containing two or three motions where one object is fixed (i.e.\ it is a part of the background), and we acquired from 6 to 10 images of size $2922 \times 2000$ with a moving camera. Fig.~\ref{fig:our_dataset} shows a sample image from each sequence. 
SIFT correspondences on such images are very noisy, as shown in Fig.~\ref{fig:mismatches}, making motion segmentation a challenging task.
In the case of the \emph{Penguin} sequence there is no motion between some frames, so pairwise segmentation was not performed. In the remaining sequences, RPA was applied to all the image pairs.


\begin{table*}[htbp]
\centering
\caption{Misclassification error [$\%$] and classified points [$\%$] for several methods on our dataset. The number of motions $d$, the number of images $n$, and the total number of image points $p$ are also reported for each sequence.
 \label{tab_real}}
\resizebox{1\textwidth}{!}{
\begin{tabular}{ l cccc c c c c c c c c c c c} 
\hline\noalign{\smallskip}
 & & & & \multicolumn{2}{c}{\ourmethod} & \multicolumn{2}{c}{Baseline} & \multicolumn{2}{c}{StableSfM + Subset \cite{XuCheongAl18}} & \multicolumn{2}{c}{QuichMatch + Subset \cite{XuCheongAl18}} & \multicolumn{2}{c}{StableSfM + RSIM \cite{JiSalzmanLi15} } & \multicolumn{2}{c}{QuichMatch + RSIM \cite{JiSalzmanLi15} } \\
 Dataset  & $d$ & $n$ & $p$ & Error & Classified & Error & Classified & Error & Classified & Error & Classified & Error & Classified  & Error & Classified  \\
\noalign{\smallskip}
\hline
\noalign{\smallskip} 
\emph{Penguin} & 2 & 6 & 5865 & \bf 0.76 & 69.17 & 0.95 & 33.95 & 32.27 & 99.59 & 41.05 & 70.11 & 41.50 & 99.59 & 41.54 & 70.11 \\
\emph{Flowers}& 2 & 6	& 7743 & \bf 1.23 & 73.65 & 2.84 & 32.70 & 8.55  & 99.50 & 8.59  & 72.59 & 16.65 & 99,50 & 14.20 & 72.59\\
\emph{Pencils}& 2 & 6	& 2982 & 3.80 & 65.33 & \bf 2.30 & 30.65 & 41.46 & 99.56 & 40.88 & 66.36 & 23.07 & 99.56 & 23.45 & 66.36 \\
\emph{Bag}      & 2 & 7	& 6114 & \bf 1.52 & 57.95 & 1.54 & 26.56 & 14.22 & 99.69 & 15.67 & 65.85 & 34.55 & 99.69 & 39.92 & 65.85\\
\emph{Bears}    & 3 & 10& 15888 & 4.82 & 73.65 & \bf 2.72 & 29.80 & 38.13 & 99.58 & 35.21 & 63.12 & 49.48 & 99.58 & 53.80 & 63.12 \\ 
\noalign{\smallskip}
\hline
\end{tabular}
}
\end{table*}

 \begin{figure*}[htbp] 
  \centering 
     \subfloat[Penguin]{
\includegraphics[width=0.19\linewidth]{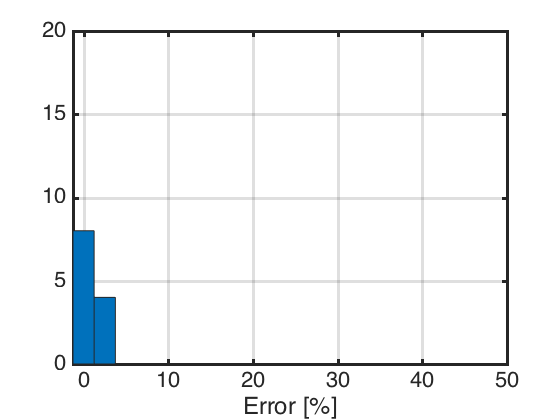}
}
   \subfloat[Flowers]{
\includegraphics[width=0.19\linewidth]{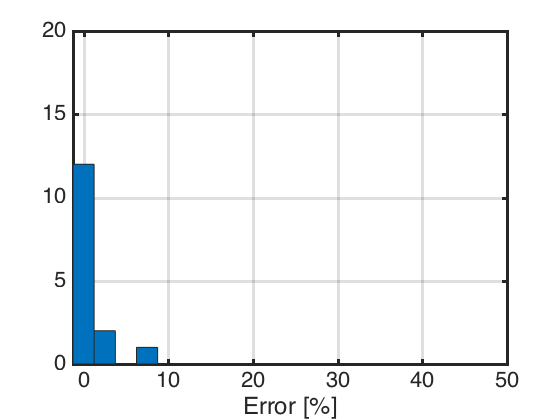}
}
   \subfloat[Pencils]{
\includegraphics[width=0.19\linewidth]{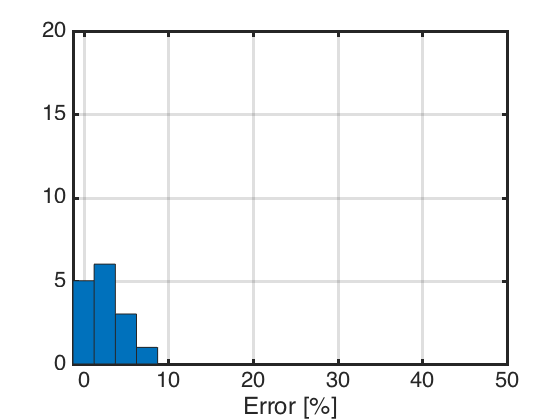}
} 
   \subfloat[Bag]{
\includegraphics[width=0.19\linewidth]{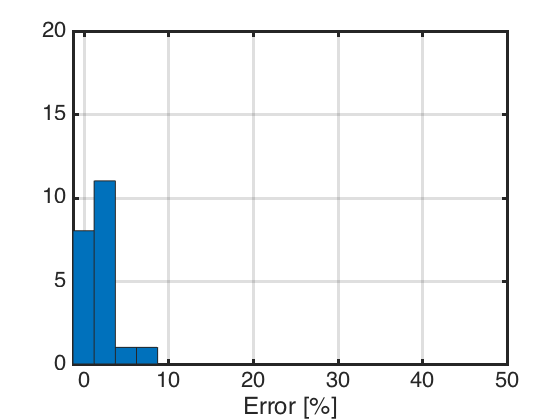}
}
   \subfloat[Bears]{
\includegraphics[width=0.19\linewidth]{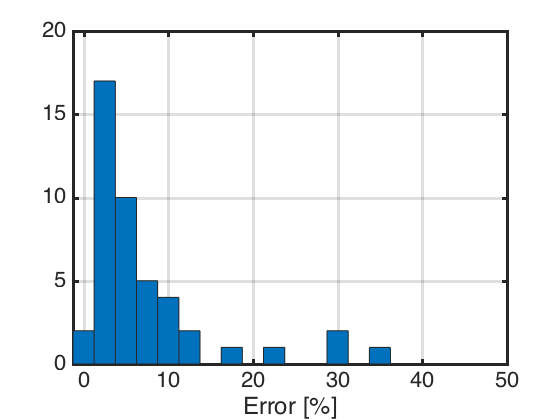}
}
\caption{
Histograms of misclassification error achieved by RPA \cite{MagriFusiello15} on all the sequences from our dataset.
Each point in the horizontal axis corresponds to a possible misclassification error in an individual image pair, and each point in the vertical axis corresponds to the number of pairs where a given error is reached.
}
\label{fig:hist_real}
\end{figure*}

\begin{figure}[htbp] 
\centering 
\includegraphics[width=0.6\linewidth]{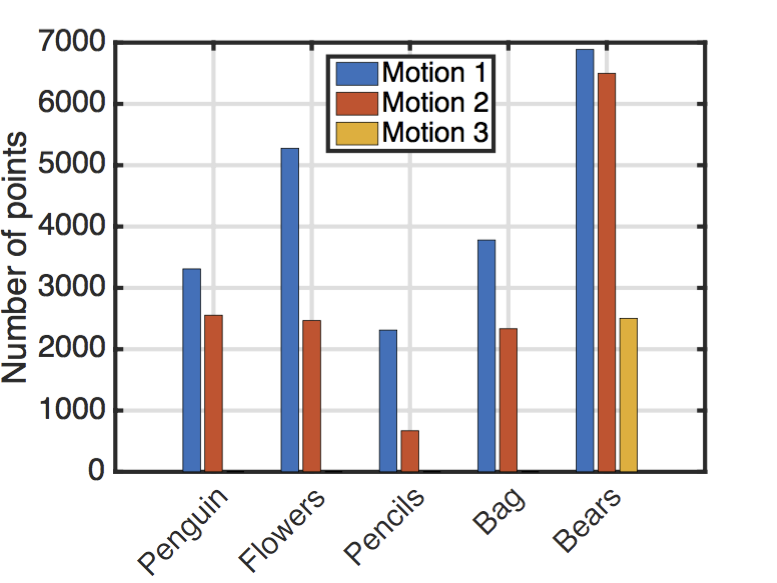}
\caption{
The number of points per motion is reported for each sequence in our dataset. 
}
\label{fig:points_per_motion}
\end{figure}

As in Sec.~\ref{sec:synthetic_experiments}, we compared \ourmethod with the baseline, which takes as input the results from pairwise segmentation, and we also considered two traditional methods, namely RSIM \cite{JiSalzmanLi15} and Subset \cite{XuCheongAl18}, where StableSfM \cite{OlssonEnqvist11} and QuichMatch \cite{TronZhouAl17} were used to compute tracks over multiple images.
In order to evaluate results quantitatively, we manually labelled points in each sequence, thus producing a ground-truth segmentation of each image, that was used to compute the misclassification error.
The number of points that undergo the same motion is reported in Fig.~\ref{fig:points_per_motion}, which gives an idea about the distribution of points in the scene for each sequence.
Results are shown in Tab.~\ref{tab_real}, which also reports the percentage of points classified by each method. 

While there are no significant differences between \ourmethod and the baseline in terms of misclassification error, the former is superior in terms of the percentage of classified points since it exploits \emph{redundant} two-frame segmentations. Both our method and the baseline -- with a misclassification error lower than $5\%$ in all the sequences -- are significantly better than Subset and RSIM. Traditional methods exhibit poor performances on our dataset since they do not deal with mismatches, confirming the outcome of the experiments on synthetic data.

\begin{figure*}[htbp] 
 \includegraphics[width=1\linewidth]{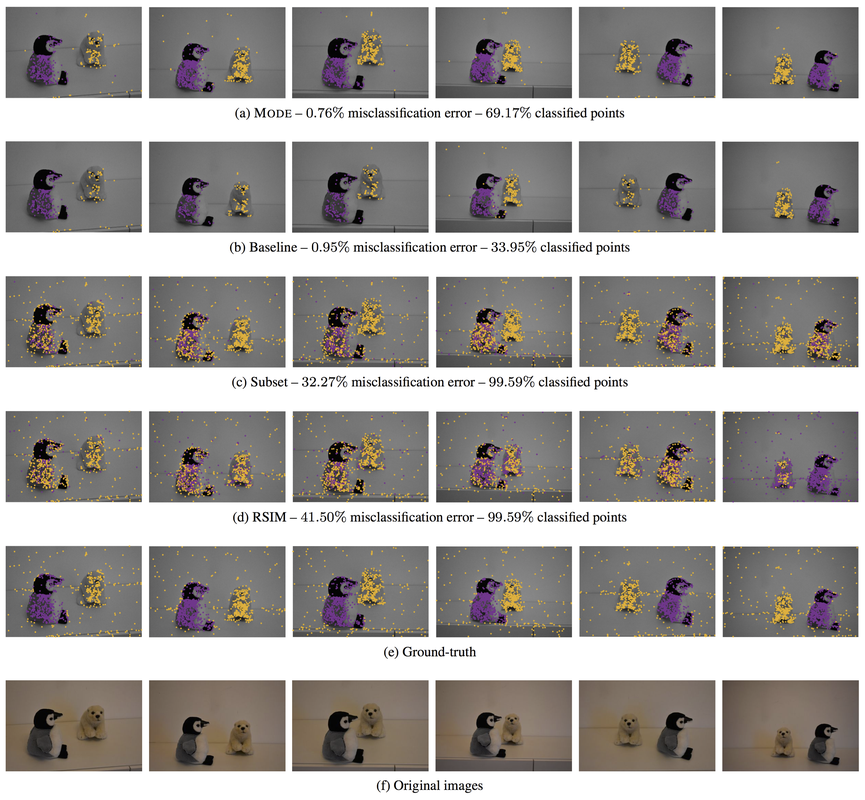}
\caption{
Segmentation results are reported for several methods on the \emph{Penguin} sequence. Images are drawn in grey-scale and points are drawn in different colors based on the membership to different motions. For better visualization, unclassified points are not drawn. Ground-truth segmentation is also reported, in addition to original (coloured) images.
}
\label{fig:penguin}
\end{figure*}

\begin{figure*}[htbp] 
  \centering 
 \includegraphics[width=1\linewidth]{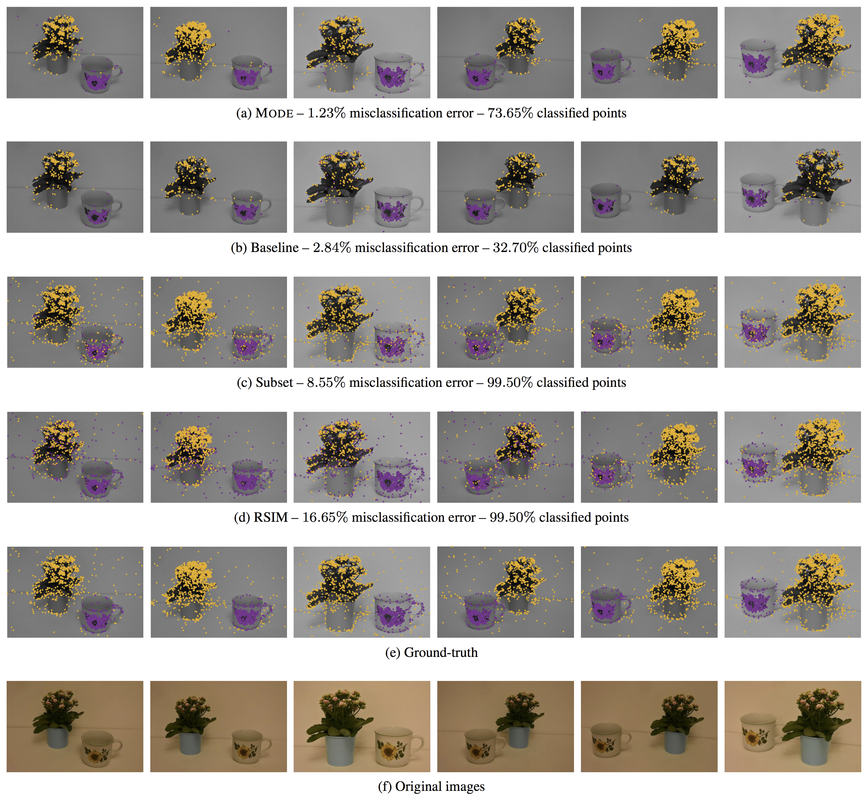}
\caption{
Segmentation results are reported for several methods on the \emph{Flowers} sequence. Images are drawn in grey-scale and points are drawn in different colors based on the membership to different motions. For better visualization, unclassified points are not drawn. Ground-truth segmentation is also reported, in addition to original (coloured) images.
}
\label{fig:flowers}
\end{figure*}

\begin{figure*}[htbp] 
  \centering 
 \includegraphics[width=1\linewidth]{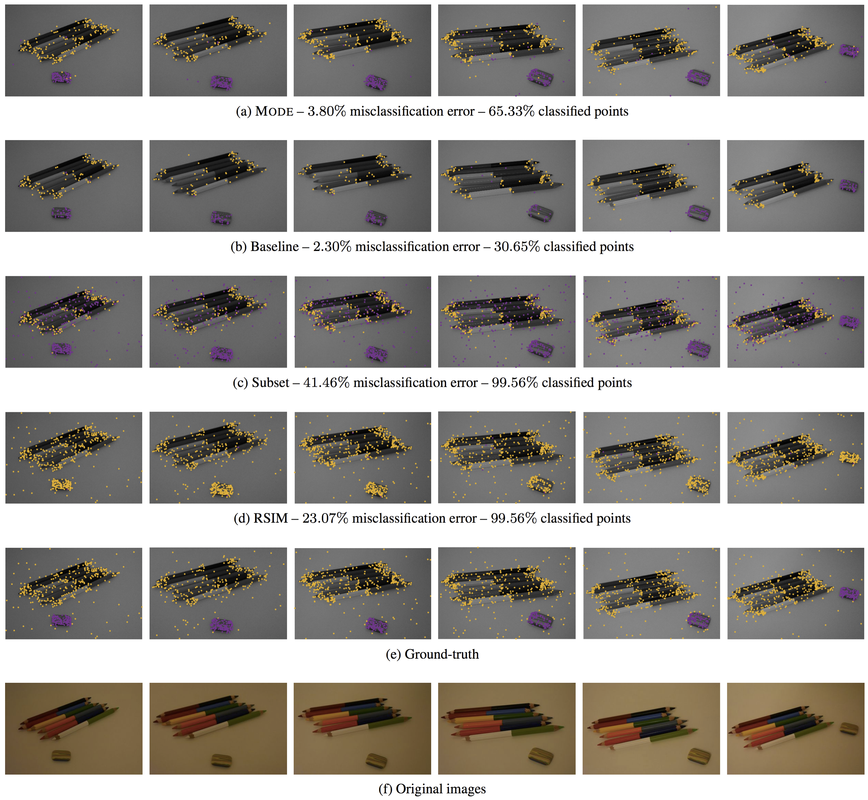}
\caption{
Segmentation results are reported for several methods on the \emph{Pencils} sequence. Images are drawn in grey-scale and points are drawn in different colors based on the membership to different motions. For better visualization, unclassified points are not drawn. Ground-truth segmentation is also reported, in addition to original (coloured) images.
}
\label{fig:pencils}
\end{figure*}

\begin{figure*}[htbp] 
  \centering 
 \includegraphics[width=1\linewidth]{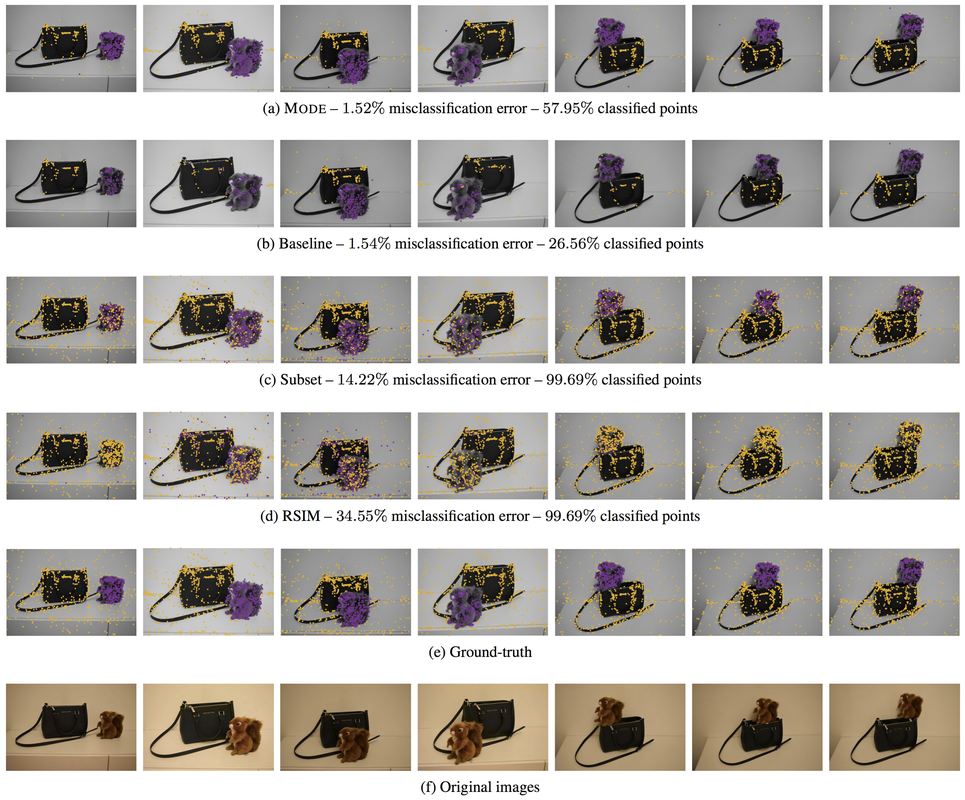}
\caption{
Segmentation results are reported for several methods on the \emph{Bag} sequence. Images are drawn in grey-scale and points are drawn in different colors based on the membership to different motions. For better visualization, unclassified points are not drawn. Ground-truth segmentation is also reported, in addition to original (coloured) images.
}
\label{fig:bag}
\end{figure*}

\begin{figure*}[htbp] 
  \centering 
 \includegraphics[width=1\linewidth]{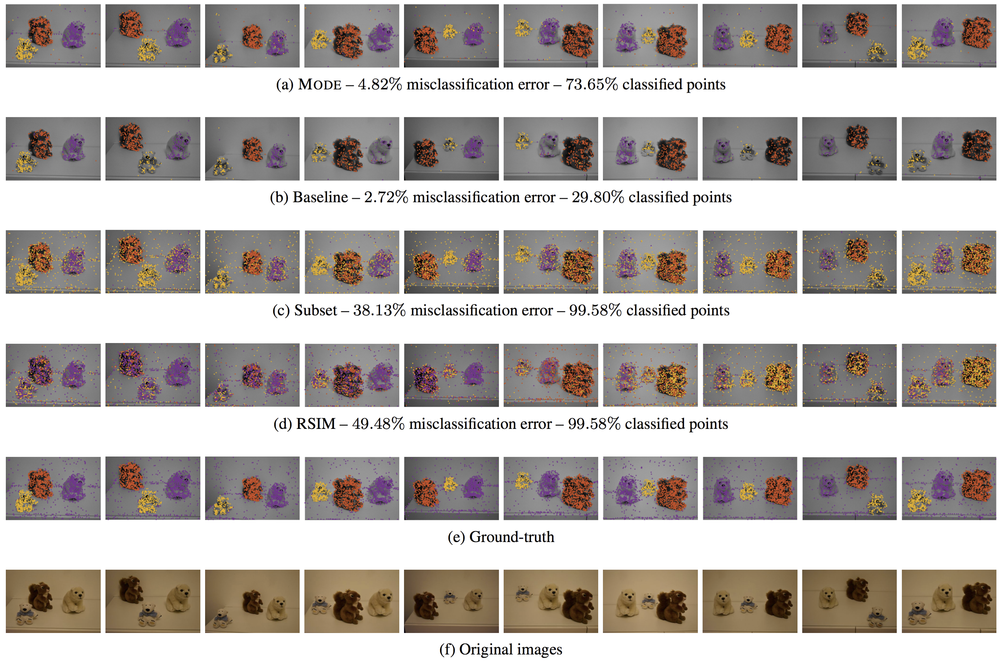}
\caption{
Segmentation results are reported for several methods on the \emph{Bears} sequence. Images are drawn in grey-scale and points are drawn in different colors based on the membership to different motions. For better visualization, unclassified points are not drawn. Ground-truth segmentation is also reported, in addition to original (coloured) images.
}
\label{fig:bears}
\end{figure*}

Figures \ref{fig:final_segmentation}, \ref{fig:penguin}, \ref{fig:flowers}, \ref{fig:pencils}, \ref{fig:bag} and \ref{fig:bears} visually represent the segmentation of image points obtained by several methods, which complement the quantitative evaluation provided in Tab.~\ref{tab_real}. Ground-truth segmentation is also shown.
Concerning the different variants of Subset \cite{XuCheongAl18} and RSIM \cite{JiSalzmanLi15}, which differ for the algorithm used for computing tracks, we report results for StableSfM \cite{OlssonEnqvist11}  only. 
Indeed, there are not significative differences between StableSfM \cite{OlssonEnqvist11} and QuichMatch \cite{TronZhouAl17} in terms of misclassification error, but the former is better in terms of amount of classified data. 
Our method returns high quality (although not perfect) segmentation in all the sequences, outperforming the baseline in terms of percentage of classified points, whereas Subset and RSIM present poor performances on our benchmark.

In order to give further insights on the behavior of our technique, we report in Fig.~\ref{fig:hist_real} the histograms of misclassification error achieved by RPA \cite{MagriFusiello15} over image pairs, similarly to Fig.~\ref{fig:histograms}.
The histograms show the effective amount of corruption in the data after performing pairwise segmentation with RPA, which is the first step of our pipeline. 
Note that the misclassification error exceeds $30\%$ in some image pairs from the \emph{Bears} sequence.
It is remarkable that our method is able to achieve a low error in this dataset (about $4.8 \%$), as reported in Tab.~\ref{tab_real}. In other words, it can effectively reduce errors in the pairwise segmentations thanks to the fact that it exploits redundant measures.

We also tested the method developed in \cite{JiLiAl14}, which does not require pairwise matches but feature locations and descriptors only. We ran the available Matlab implementation of~\cite{JiLiAl14} on \emph{Pencils} sequence. It did not return any solution after several hours of computation due to ``out of memory'' error. We conclude that it is not yet a practical approach to motion segmentation on the scenarios considered in our paper.

\subsubsection{Outdoor scenes}
\begin{figure*}[htbp] 
  \centering 
 \includegraphics[width=1\linewidth]{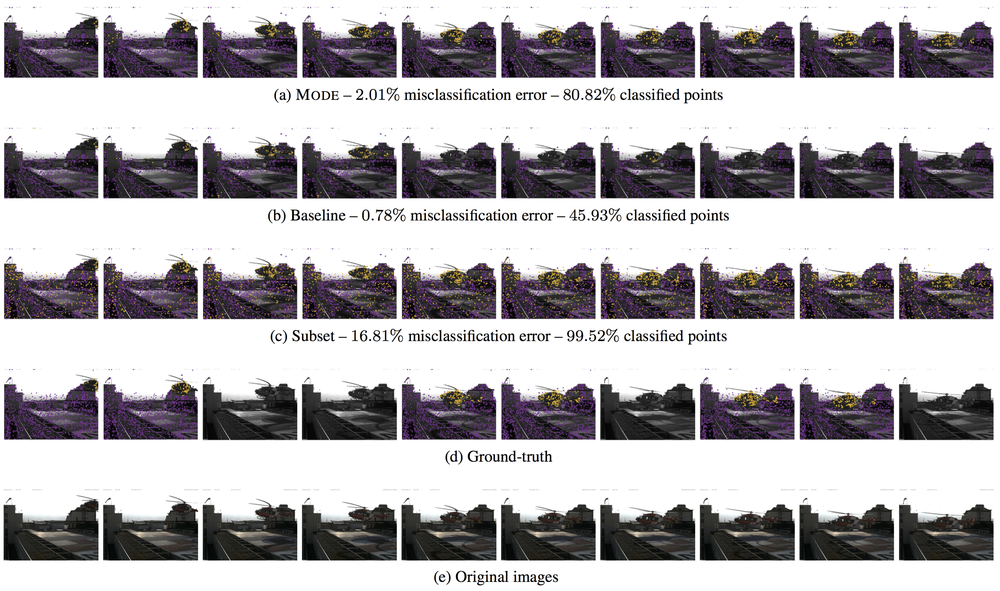}
\caption{
Segmentation results are reported for several methods on the \emph{helicopter} sequence \cite{DragonOstermannAl13}. Images are drawn in grey-scale and points are drawn in different colors based on the membership to different motions. For better visualization, unclassified points are not drawn. Ground-truth segmentation is reported only for those images for which ground-truth pixel-wise annotation is provided.
Original (coloured) images are also reported.
}
\label{fig:heli}
\end{figure*}

\begin{figure*}[htbp] 
  \centering 
  \includegraphics[width=1\linewidth]{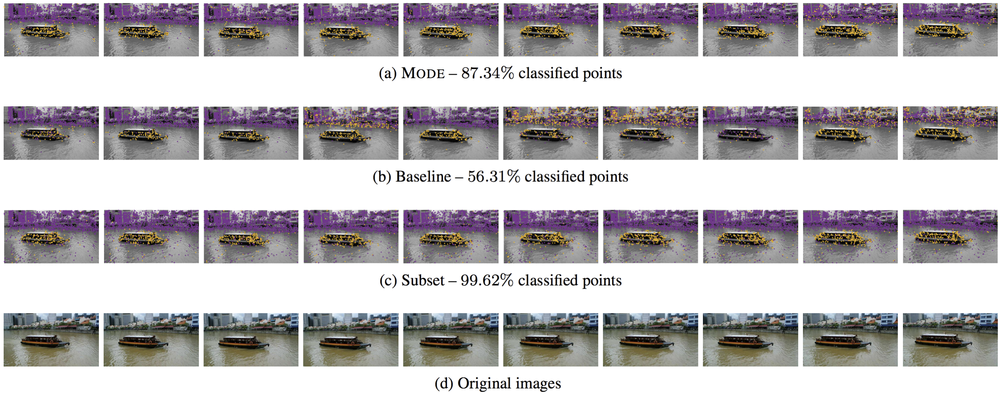}
  \caption{
Segmentation results are reported for several methods on the \emph{boat} sequence \cite{LiGuoAl13}. Images are drawn in grey-scale and points are drawn in different colors based on the membership to different motions. For better visualization, unclassified points are not drawn.
Original (coloured) images are also reported.
}
\label{fig:boat}
\end{figure*}

\begin{figure*}[htbp] 
  \centering 
 \includegraphics[width=1\linewidth]{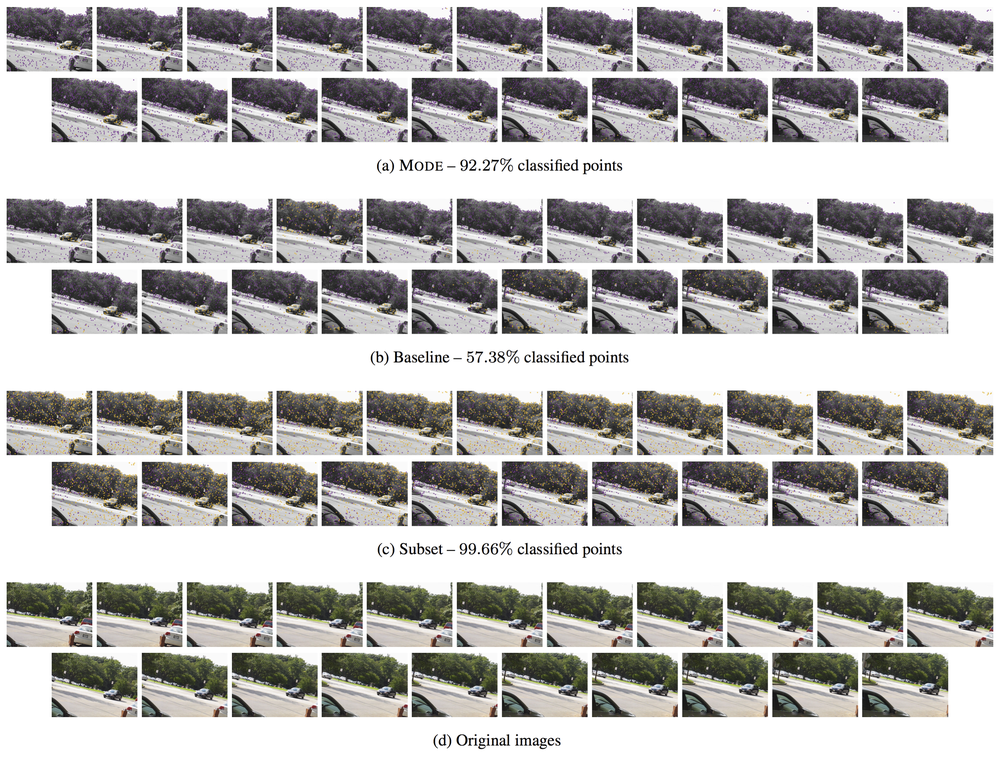}
\caption{
Segmentation results are reported for several methods on the \emph{cars7} sequence \cite{TronVidal07}. Images are drawn in grey-scale and points are drawn in different colors based on the membership to different motions. For better visualization, unclassified points are not drawn.
Original (coloured) images are also reported.
}
\label{fig:cars7}
\end{figure*}

\begin{figure*}[htbp] 
  \centering 
 \includegraphics[width=1\linewidth]{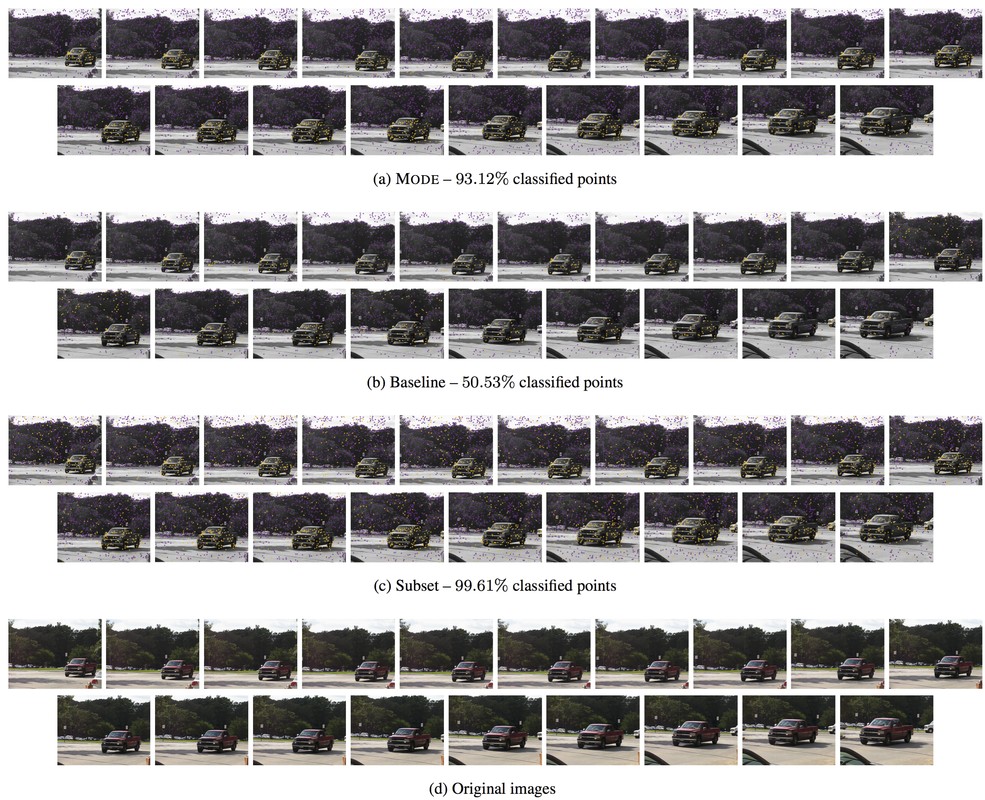}
\caption{
Segmentation results are reported for several methods on the \emph{cars8} sequence \cite{TronVidal07}. Images are drawn in grey-scale and points are drawn in different colors based on the membership to different motions. For better visualization, unclassified points are not drawn.
Original (coloured) images are also reported.
}
\label{fig:cars8}
\end{figure*}

\noindent To study a more realistic scenario, we considered four outdoor scenes, namely \emph{helicopter} \cite{DragonOstermannAl13}, \emph{boat} \cite{LiGuoAl13}, \emph{cars7} \cite{TronVidal07} and \emph{cars8} \cite{TronVidal07}, which are shown in in Fig.~\ref{fig:heli}, \ref{fig:boat}, \ref{fig:cars7} and \ref{fig:cars8}. A subset of the images was chosen for each sequence in order to ensure enough motion between consecutive frames. The properties of each dataset are presented in Tab.~\ref{tab_outdoor}, which also reports the percentage of points classified by \ourmethod, the baseline and Subset \cite{XuCheongAl18} combined with StableSfM \cite{OlssonEnqvist11}. The latter provided the best results among all possible combinations of traditional segmentation methods and tracking algorithms.
%
%
%
In the case of the \emph{helicopter} sequence, a subset of the images has ground-truth pixel-wise annotation, which was used to compute the misclassification error (see Tab.~\ref{tab_outdoor}). For the remaining sequences, no ground-truth is available, so only qualitative evaluation can be provided, which is reported in Fig.~\ref{fig:heli}, \ref{fig:boat}, \ref{fig:cars7} and \ref{fig:cars8}.

Results show that our solution is of good quality in all the images, outperforming the baseline in terms of amount of classified data. 
%
The poor performance of the baseline on some images gives an idea about how noisy the individual pairwise segmentations are. Our method is able to reduce such errors thanks to the fact that it  exploits redundant measures.
There are no significant differences between Subset and \ourmethod in the \emph{boat} sequence, which, however, is a simple scene for matching due to slow motion. In the \emph{helicopter}, \emph{cars7} and \emph{cars8} sequences,  Subset produces useless results.
Table \ref{tab_outdoor} shows that our method is significantly better than Subset in terms of segmentation accuracy on the \emph{helicopter} scene.
Although the baseline achieves the lowest error, it must be noted that it does not provide a useful solution to segmentation since it classifies less than $50\%$ of the points. This can also be seen in Fig.~\ref{fig:heli} where the baseline is not able to classify any point in the moving object in 5 out of 10 images.

\begin{table}[htbp]
\centering
\caption{Misclassification error [$\%$] and classified points [$\%$] for several methods on outdoor scenes. The number of motions $d$, the number of images $n$, and the total number of image points $p$ are also reported for each sequence.
 \label{tab_outdoor}}
\resizebox{1\columnwidth}{!}{
\begin{tabular}{ l c c c c c c c c c c c c c c c} 
\hline\noalign{\smallskip}
 & & & & \multicolumn{2}{c}{\ourmethod} & \multicolumn{2}{c}{Baseline} & \multicolumn{2}{c}{StableSfM + Subset \cite{XuCheongAl18}} \\
 Dataset  & $d$ & $n$ & $p$ & Error & Classified & Error & Classified & Error & Classified  \\
\noalign{\smallskip}
\hline
\noalign{\smallskip} 
\emph{helicopter} \cite{DragonOstermannAl13} & 2 & 10 & 17139 & 2.01 & 80.82 & \bf 0.78 & 45.93 & 16.81 & 99.52 \\
\emph{boat}  \cite{LiGuoAl13}   & 2 & 10 & 21183 & -- & 87.34 & -- & 56.31 & -- & 99.62   \\
\emph{cars7} \cite{TronVidal07}  & 2 & 21 & 16602 & -- & 92.27 & -- & 57.38 & -- & 99.66  \\
\emph{cars8} \cite{TronVidal07}  & 2 & 19 & 13438 & -- & 93.12 & -- & 50.53 & -- & 99.61  \\
\noalign{\smallskip}
\hline
\end{tabular}
}
\end{table}

\section{Conclusion}
\label{sec:conclusion}
\noindent We presented a new solution to the motion segmentation where the problem is split in two steps. First, a segmentation is performed independently on pairs of images. Then, the partial/local results are combined to segment points in all the images. This general framework -- combined with a robust solution to two-frame motion segmentation (e.g.~RPA \cite{MagriFusiello15}) -- handles realistic situations such as the presence of mismatches that have been overlooked so far in previous motion segmentation work. Our approach does not require tracks as input but only pairwise correspondences. Thus it could be exploited to build tracks that are aware of segmentation, which constitutes the foundation of a multibody structure from motion pipeline. Future research will explore this direction. 

\paragraph{Acknowledgements.}

The authors would like to thank Luca Magri for his guidance through the Matlab code of RPA \cite{MagriFusiello15} and Stanislav Steidl for his help with the experiments.
This work was supported by the European Regional Development Fund under the project IMPACT (reg.~no CZ$.02.1.01/0.0/0.0/15\_003/0000468$).

{\small
\bibliographystyle{ieee}
\bibliography{arxiv}

\begin{thebibliography}{10}\itemsep=-1pt

\bibitem{ArrigoniFusielloAl15siims}
F.~Arrigoni, B.~Rossi, and A.~Fusiello.
\newblock Spectral synchronization of multiple views in {SE(3)}.
\newblock {\em SIAM Journal on Imaging Sciences}, 9(4):1963 -- 1990, 2016.

\bibitem{BarathMatas18}
D.~Barath and J.~Matas.
\newblock Multi-class model fitting by energy minimization and mode-seeking.
\newblock In {\em Proceedings of the European Conference on Computer Vision},
  pages 229--245. Springer International Publishing, 2018.

\bibitem{BoydParikh11}
S.~Boyd, N.~Parikh, E.~Chu, B.~Peleato, and J.~Eckstein.
\newblock Distributed optimization and statistical learning via the alternating
  direction method of multipliers.
\newblock {\em Foundations and Trends in Machine Learning}, 3(1):1--122, Jan.
  2011.

\bibitem{ChatterjeeGovindu17}
A.~Chatterjee and V.~M. Govindu.
\newblock Robust relative rotation averaging.
\newblock {\em IEEE Transactions on Pattern Analysis and Machine Intelligence},
  2017.

\bibitem{ChinSuterAl10}
T.-J. Chin, D.~Suter, and H.~Wang.
\newblock Multi-structure model selection via kernel optimisation.
\newblock In {\em Proceedings of the IEEE Conference on Computer Vision and
  Pattern Recognition}, pages 3586--3593, 2010.

\bibitem{CortesMohriAl09}
C.~Cortes, M.~Mohri, and A.~Rostamizadeh.
\newblock Learning non-linear combinations of kernels.
\newblock In {\em Neural Information Processing Systems}, pages 396--404. 2009.

\bibitem{CosteiraKanade98}
J.~P. Costeira and T.~Kanade.
\newblock A multibody factorization method for independently moving objects.
\newblock {\em International Journal of Computer Vision}, 29(3):159--179, 1998.

\bibitem{DelongGorelickAl12}
A.~Delong, L.~Gorelick, O.~Veksler, and Y.~Boykov.
\newblock Minimizing energies with hierarchical costs.
\newblock {\em International Journal of Computer Vision}, 100(1):38--58, 2012.

\bibitem{DelongOsokinAl12}
A.~Delong, A.~Osokin, H.~N. Isack, and Y.~Boykov.
\newblock Fast approximate energy minimization with label costs.
\newblock {\em International Journal of Computer Vision}, 96(1):1--27, 2012.

\bibitem{DragonOstermannAl13}
R.~Dragon, J.~Ostermann, and L.~Van~Gool.
\newblock Robust realtime motion-split-and-merge for motion segmentation.
\newblock In {\em German Conference on Pattern Recognition}, pages 425--434.
  Springer Berlin Heidelberg, 2013.

\bibitem{ElhamifarVidal13}
E.~Elhamifar and R.~Vidal.
\newblock Sparse subspace clustering: Algorithm, theory, and applications.
\newblock {\em IEEE Transactions on Pattern Analysis and Machine Intelligence},
  35(11):2765--2781, 2013.

\bibitem{EssTobiasAl09}
A.~Ess, T.~Mueller, H.~Grabner, and L.~V. Gool.
\newblock Segmentation-based urban traffic scene understanding.
\newblock In {\em British Machine Vision Conference}, 2009.

\bibitem{Gear98}
C.~W. Gear.
\newblock Multibody grouping from motion images.
\newblock {\em International Journal of Computer Vision}, 29(2):133--150, 1998.

\bibitem{HartleyAftabAl11}
R.~Hartley, K.~Aftab, and J.~Trumpf.
\newblock L1 rotation averaging using the {Weiszfeld} algorithm.
\newblock {\em Proceedings of the IEEE Conference on Computer Vision and
  Pattern Recognition}, pages 3041--3048, 2011.

\bibitem{IsackBoykov12}
H.~Isack and Y.~Boykov.
\newblock Energy-based geometric multi-model fitting.
\newblock {\em International Journal of Computer Vision}, 97(2):123--147, 2012.

\bibitem{JiLiAl14}
P.~Ji, H.~Li, M.~Salzmann, and Y.~Dai.
\newblock Robust motion segmentation with unknown correspondences.
\newblock In {\em Proceedings of the European Conference on Computer Vision},
  pages 204--219. Springer International Publishing, 2014.

\bibitem{JiSalzmanLi15}
P.~Ji, M.~Salzmann, and H.~Li.
\newblock Shape interaction matrix revisited and robustified: Efficient
  subspace clustering with corrupted and incomplete data.
\newblock In {\em Proceedings of the International Conference on Computer
  Vision}, pages 4687--4695, 2015.

\bibitem{KimKim03}
J.~B. Kim and H.~J. Kim.
\newblock Efficient region-based motion segmentation for a video monitoring
  system.
\newblock {\em Pattern Recognition Letters}, 24(1):113 -- 128, 2003.

\bibitem{Kuhn55}
H.~W. Kuhn.
\newblock The {Hungarian} method for the assignment problem.
\newblock {\em Naval Research Logistics Quarterly 2}, 2:83 -- 97, 1955.

\bibitem{KumarRaiAl11}
A.~Kumar, P.~Rai, and H.~Daume.
\newblock Co-regularized multi-view spectral clustering.
\newblock In {\em Neural Information Processing Systems}, pages 1413--1421.
  2011.

\bibitem{LaiWangAl17}
T.~Lai, H.~Wang, Y.~Yan, T.~Chin, and W.~Zhao.
\newblock Motion segmentation via a sparsity constraint.
\newblock {\em IEEE Transactions on Intelligent Transportation Systems},
  18(4):973--983, 2017.

\bibitem{LiVidal15}
C.-G. Li and R.~Vidal.
\newblock Structured sparse subspace clustering: A unified optimization
  framework.
\newblock In {\em Proceedings of the IEEE Conference on Computer Vision and
  Pattern Recognition}, pages 277--286, 2015.

\bibitem{LiKallemAl07}
T.~Li, V.~Kallem, D.~Singaraju, and R.~Vidal.
\newblock Projective factorization of multiple rigid-body motions.
\newblock In {\em Proceedings of the IEEE Conference on Computer Vision and
  Pattern Recognition}, pages 1--6, 2007.

\bibitem{LiGuoAl13}
Z.~Li, J.~Guo, L.~Cheong, and S.~Z. Zhou.
\newblock Perspective motion segmentation via collaborative clustering.
\newblock In {\em Proceedings of the International Conference on Computer
  Vision}, pages 1369--1376, 2013.

\bibitem{LiuLinAl13}
G.~Liu, Z.~Lin, S.~Yan, J.~Sun, Y.~Yu, and Y.~Ma.
\newblock Robust recovery of subspace structures by low-rank representation.
\newblock {\em IEEE Transactions on Pattern Analysis and Machine Intelligence},
  pages 171--184, 2013.

\bibitem{Lowe04}
D.~G. Lowe.
\newblock Distinctive image features from scale-invariant keypoints.
\newblock {\em International Journal of Computer Vision}, 60(2):91--110, 2004.

\bibitem{MagriFusiello14}
L.~Magri and A.~Fusiello.
\newblock {T-Linkage}: A continuous relaxation of {J-Linkage} for multi-model
  fitting.
\newblock In {\em Proceedings of the IEEE Conference on Computer Vision and
  Pattern Recognition}, pages 3954--3961, June 2014.

\bibitem{MagriFusiello15}
L.~Magri and A.~Fusiello.
\newblock Robust multiple model fitting with preference analysis and low-rank
  approximation.
\newblock In {\em Proceedings of the British Machine Vision Conference}, pages
  20.1--20.12. BMVA Press, September 2015.

\bibitem{MagriFusiello16}
L.~Magri and A.~Fusiello.
\newblock Multiple models fitting as a set coverage problem.
\newblock In {\em Proceedings of the IEEE Conference on Computer Vision and
  Pattern Recognition}, pages 3318--3326, June 2016.

\bibitem{OlssonEnqvist11}
C.~Olsson and O.~Enqvist.
\newblock Stable structure from motion for unordered image collections.
\newblock In {\em Proceedings of the 17th Scandinavian conference on Image
  analysis (SCIA'11)}, pages 524--535. Springer-Verlag, 2011.

\bibitem{OzdenSchindlerAl10}
K.~E. Ozden, K.~Schindler, and L.~Van~Gool.
\newblock Multibody structure-from-motion in practice.
\newblock {\em IEEE Transactions on Pattern Analysis and Machine Intelligence},
  32(6):1134--1141, 2010.

\bibitem{OzyesilVoroninskiAl17}
O.~Ozyesil, V.~Voroninski, R.~Basri, and A.~Singer.
\newblock A survey of structure from motion.
\newblock {\em Acta Numerica}, 26:305 -- 364, 2017.

\bibitem{PachauriKondorAl13}
D.~Pachauri, R.~Kondor, and V.~Singh.
\newblock Solving the multi-way matching problem by permutation
  synchronization.
\newblock In {\em Advances in Neural Information Processing Systems 26}, pages
  1860--1868. Curran Associates, Inc., 2013.

\bibitem{PhamChinAl14}
T.-T. Pham, T.-J. Chin, J.~Yu, and D.~Suter.
\newblock The random cluster model for robust geometric fitting.
\newblock {\em IEEE Transactions on Pattern Analysis and Machine Intelligence},
  36(8):1658--1671, 2014.

\bibitem{QianChellappaAl05}
G.~Qian, R.~Chellappa, and Q.~Zheng.
\newblock Bayesian algorithms for simultaneous structure from motion estimation
  of multiple independently moving objects.
\newblock {\em IEEE Transactions on Image Processing}, 14(1):94--109, 2005.

\bibitem{RaoTronAl10}
S.~Rao, R.~Tron, R.~Vidal, and Y.~Ma.
\newblock Motion segmentation in the presence of outlying, incomplete, or
  corrupted trajectories.
\newblock {\em Pattern Analysis and Machine Intelligence}, 32(10):1832--1845,
  2010.

\bibitem{RubinoDelbueAl18}
C.~Rubino, A.~{Del Bue}, and T.-J. Chin.
\newblock Practical motion segmentation for urban street view scenes.
\newblock In {\em Proceedings of the IEEE International Conference on Robotics
  and Automation}, 2018.

\bibitem{SabzevariScaramuzza14}
R.~Sabzevari and D.~Scaramuzza.
\newblock Monocular simultaneous multi-body motion segmentation and
  reconstruction from perspective views.
\newblock In {\em Proceedings of the IEEE International Conference on Robotics
  and Automation}, pages 23--30, 2014.

\bibitem{SabzevariScaramuzza16}
R.~Sabzevari and D.~Scaramuzza.
\newblock Multi-body motion estimation from monocular vehicle-mounted cameras.
\newblock {\em IEEE Transactions on Robotics}, 32(3):638--651, 2016.

\bibitem{SaputraMarkhamAl18}
M.~R.~U. Saputra, A.~Markham, and N.~Trigoni.
\newblock Visual {SLAM} and structure from motion in dynamic environments: A
  survey.
\newblock {\em ACM Computing Surveys}, 51(2):37:1--37:36, 2018.

\bibitem{SchindlerSuterAl08}
K.~Schindler, D.~Suter, and H.~Wang.
\newblock A model-selection framework for multibody structure-and-motion of
  image sequences.
\newblock {\em International Journal of Computer Vision}, 79(2):159--177, 2008.

\bibitem{Singer11}
A.~Singer.
\newblock Angular synchronization by eigenvectors and semidefinite programming.
\newblock {\em Applied and Computational Harmonic Analysis}, 30(1):20 -- 36,
  2011.

\bibitem{ThakoorGaoAl10}
N.~Thakoor, J.~Gao, and V.~Devarajan.
\newblock Multibody structure-and-motion segmentation by branch-and-bound model
  selection.
\newblock {\em IEEE Transactions on Image Processing}, 19(6):1393--1402, 2010.

\bibitem{ToldoFusiello08}
R.~Toldo and A.~Fusiello.
\newblock Robust multiple structures estimation with {J-Linkage}.
\newblock In {\em Proceedings of the European Conference on Computer Vision},
  pages 537--547, 2008.

\bibitem{Torr98}
P.~H.~S. Torr.
\newblock Geometric motion segmentation and model selection.
\newblock {\em Philosophical Transactions of the Royal Society of London A:
  Mathematical, Physical and Engineering Sciences}, 356(1740):1321--1340, 1998.

\bibitem{TorselloRodolaAl11}
A.~Torsello, E.~Rodol\`a, and A.~Albarelli.
\newblock Multiview registration via graph diffusion of dual quaternions.
\newblock In {\em Proceedings of the IEEE Conference on Computer Vision and
  Pattern Recognition}, pages 2441 -- 2448, 2011.

\bibitem{TronVidal07}
R.~Tron and R.~Vidal.
\newblock A benchmark for the comparison of 3-d motion segmentation algorithms.
\newblock In {\em Proceedings of the IEEE Conference on Computer Vision and
  Pattern Recognition}, pages 1--8. IEEE, 2007.

\bibitem{TronZhouAl17}
R.~{Tron}, X.~{Zhou}, C.~{Esteves}, and K.~{Daniilidis}.
\newblock Fast multi-image matching via density-based clustering.
\newblock In {\em Proceedings of the International Conference on Computer
  Vision}, pages 4077--4086, 2017.

\bibitem{VidalHartley08}
R.~Vidal and R.~Hartley.
\newblock Three-view multibody structure from motion.
\newblock {\em IEEE Transactions on Pattern Analysis and Machine Intelligence},
  30(2):214--227, 2008.

\bibitem{VidalMaAl05}
R.~Vidal, Y.~Ma, and S.~Sastry.
\newblock Generalized principal component analysis (gpca).
\newblock {\em IEEE Transactions on Pattern Analysis and Machine Intelligence},
  27(12):1945--1959, 2005.

\bibitem{VidalMaAl06}
R.~Vidal, Y.~Ma, S.~Soatto, and S.~Sastry.
\newblock Two-view multibody structure from motion.
\newblock {\em International Journal of Computer Vision}, 68(1):7--25, 2006.

\bibitem{VidalTronAl08}
R.~Vidal, R.~Tron, and R.~Hartley.
\newblock Multiframe motion segmentation with missing data using
  powerfactorization and {GPCA}.
\newblock {\em International Journal of Computer Vision}, 79(1):85--105, 2008.

\bibitem{VincentLaganiere01}
E.~Vincent and R.~Laganiere.
\newblock Detecting planar homographies in an image pair.
\newblock In {\em International Symposium on Image and Signal Processing and
  Analysis}, pages 182--187, 2001.

\bibitem{Von-Luxburg07}
U.~Von~Luxburg.
\newblock A tutorial on spectral clustering.
\newblock {\em Statistics and computing}, 17(4):395--416, 2007.

\bibitem{WangQianAl14}
X.~Wang, B.~Qian, and I.~Davidson.
\newblock On constrained spectral clustering and its applications.
\newblock {\em Data Mining and Knowledge Discovery}, 28(1):1--30, 2014.

\bibitem{WeinlandRonfardAl11}
D.~Weinland, R.~Ronfard, and E.~Boyer.
\newblock A survey of vision-based methods for action representation,
  segmentation and recognition.
\newblock {\em Computer Vision and Image Understanding}, 115(2):224 -- 241,
  2011.

\bibitem{XuOjaAl90}
L.~Xu, E.~Oja, and P.~Kultanen.
\newblock A new curve detection method: randomized {H}ough transform ({RHT}).
\newblock {\em Pattern Recognition Letters}, 11(5):331--338, 1990.

\bibitem{XuCheongAl18}
X.~Xu, L.-F. Cheong, and Z.~Li.
\newblock Motion segmentation by exploiting complementary geometric models.
\newblock In {\em Proceedings of the IEEE Conference on Computer Vision and
  Pattern Recognition}, 2018.

\bibitem{YanPollefeys06}
J.~Yan and M.~Pollefeys.
\newblock A general framework for motion segmentation: Independent,
  articulated, rigid, non-rigid, degenerate and nondegenerate.
\newblock In {\em Proceedings of the European Conference on Computer Vision},
  pages 94--106, 2006.

\bibitem{ZappellaDelbueAl13}
L.~Zappella, A.~D. Bue, X.~Llad\'o, and J.~Salvi.
\newblock Joint estimation of segmentation and structure from motion.
\newblock {\em Computer Vision and Image Understanding}, 117(2):113 -- 129,
  2013.

\bibitem{ZhangKosecka06}
W.~Zhang and J.~Koseck{\'a}.
\newblock Nonparametric estimation of multiple structures with outliers.
\newblock In {\em Workshop on Dynamic Vision, European Conference on Computer
  Vision 2006}, volume 4358 of {\em Lecture Notes in Computer Science}, pages
  60--74. Springer, 2006.

\bibitem{ZulianiKenneyAl05}
M.~Zuliani, C.~S. Kenney, and B.~S. Manjunath.
\newblock The multi{RANSAC} algorithm and its application to detect planar
  homographies.
\newblock In {\em Proceedings of the IEEE International Conference on Image
  Processing}, pages III--153--6, September 11-14 2005.

\end{thebibliography}
}

\end{document}